\documentclass[11pt]{article}
\usepackage{amssymb,amsmath,amsthm,mathrsfs} %this package contains all the special AMS fonts etc.
\usepackage{colordvi,color,pspicture}
\usepackage{graphicx}        % main graphic packages
\usepackage{epstopdf}
\usepackage{algorithm}       % to write the algorithms
\usepackage{algorithmic}
\usepackage{rotating}        % sideways command
\usepackage{multirow}        % for multirows in tables
\usepackage{natbib}          % references
\usepackage{url}
\usepackage[font=footnotesize,labelfont=bf]{caption} % for small captions
\newcommand{\Vol}{\text{Vol}}

\newtheoremstyle{dotless}{}{}{\itshape}{}{\bfseries}{}{ }{}
\theoremstyle{dotless}

\theoremstyle{definition}

\DeclareMathOperator*{\argmin}{\arg\!\min}
\DeclareMathOperator*{\argmax}{\arg\!\max}

\begin{document}
\title{Classification of Imbalanced Data with a Geometric Digraph Family}
\author{Artur Manukyan \& Elvan Ceyhan\thanks{Department of Statistics, North Carolina State University, Rayleigh, NC, USA.}}

\date{\today}

\maketitle
%\pagenumbering{roman} \setcounter{page}{1}
%\tableofcontents
%\newpage
\pagenumbering{arabic} \setcounter{page}{1}

\begin{abstract}%
\noindent We use a geometric digraph family called class cover catch digraphs (CCCDs) to tackle the class imbalance problem in statistical classification. CCCDs provide graph theoretic solutions to the class cover problem and have been employed in classification. We assess the classification performance of CCCD classifiers by extensive Monte Carlo simulations, comparing them with other classifiers commonly used in the literature. In particular, we show that CCCD classifiers perform relatively well when one class is more frequent than the other in a two-class setting, an example of the \emph{class imbalance problem}. We also point out the relationship between class imbalance and class overlapping problems, and their influence on the performance of CCCD classifiers and other classification methods as well as some state-of-the-art algorithms which are robust to class imbalance by construction.
Experiments on both simulated and real data sets indicate that CCCD classifiers are robust to the class imbalance problem. CCCDs substantially undersample from the majority class while preserving the information on the discarded points during the undersampling process.
Many state-of-the-art methods, however, keep this information by means of ensemble classifiers,
but CCCDs yield only a single classifier with the same property,
making it both appealing and fast.
\end{abstract}

% keywords
{\small {\it Keywords:} class cover catch digraphs, class cover problem, class imbalance problem, class overlapping problem, digraph, domination, graph, prototype selection, support estimation

\vspace{.25 in}

$^*$corresponding author.\\
\indent {\it e-mail:} elvanceyhan@gmail.com (E.~Ceyhan)}

\section{Introduction}

Class imbalance problem has recently become a topic of extensive research. In a two-class setting, imbalance in class(es) occurs when one class is represented by far more observations (points) than the other class(es) in the data set (see, e.g., \cite{chawla2004} and \cite{lopez2013}).
Class imbalance problem is observed in many areas such as medicine, fraud detection and education. Some examples are clinical trials in which only 5\% of patients in the data set have a certain disease, such as cancer \citep{mazurowski2008}; detecting fraudulent customers where most individuals are law-abiding in insurance, credit card and telecommunications industries \citep{phua2004}; and archives of college students where mostly the ones who have fair results are kept \citep{thai2009}. In these and many other real life cases, majority class (i.e., the class with larger size) confounds the classifier performance by hindering the detection of subjects from the minority class (i.e., the class with fewer points).

The classification methods in machine learning usually suffer from the imbalance of class sizes in the data sets because most of these methods work on the assumption that class sizes are balanced \citep{japkowicz2002}.
For example, the commonly used $k$-nearest neighbor ($k$-NN) classification algorithm is highly influenced by the class imbalance problem.
In the $k$-NN approach, a new point is labeled  with the label of the
most frequent class among its first $k$ nearest neighbors \citep{fix1989,cover1967}. As a result, in a two-class setting where one class substantially outnumbers the other,
a point is more likely to be classified as the majority class by the $k$-NN classifier. In literature, sensitivity of $k$-NN classifier to the class imbalance problem and some solutions on choosing the appropriate $k$ have been discussed in cases of imbalanced classes \citep[see][]{mani2003,garcia2008,hand2003}. Decision trees and support vector machines (SVM) are also some of the well known classifiers that are sensitive to the class imbalance in a data set \citep{japkowicz2002,tang2009}. SVMs are among the most commonly used algorithms in the machine learning literature due to their well understood theory and high performance among popular algorithms \citep{wu2008,fernandez2014}, but these methods have been demonstrated to be inefficient against highly imbalanced data sets, although SVMs are still robust to moderately imbalanced data sets \citep{akbani2004,raskutti2004}.

We approach the classification of imbalanced data sets with methods that solve class cover problem (CCP), where the goal is to find a region that encapsulates all the members of the class of interest (i.e., target class).
This particular region can be viewed as a \emph{cover}; hence the name \emph{class cover} \citep{cannon2004}. This problem is closely related to another problem in statistics, namely \emph{support estimation}: estimating the support of a particular random variable defined in a measurable space \citep{scholkopf2001}. Here, each cover can be viewed as estimates of its associated class support.
\citet{priebe2001} introduced the class cover catch digraphs (CCCD) to find graph theoretic solutions to the CCP problem, and provided some results on the minimum dominating sets and the distribution of the domination number of such digraphs for one dimensional data. \citet{priebe2003} applied CCCDs on classification and showed that approximate minimum dominating sets of CCCDs (which were obtained by a greedy algorithm) and radii of the covering balls can be used to establish efficient classifiers. Moreover, \citet{devinney2002} defined random walk CCCDs (RW-CCCDs) where balls of class covers are more relaxed compared to previously introduced so called pure-CCCDs (P-CCCDs). In P-CCCDs, no member of the non-target class is covered,
but RW-CCCDs allow some points of non-target class to be covered by the cover of the target class;
and some target class points may also be uncovered in the process.
Hence, RW-CCCDs may potentially avoid overfitting. CCCDs have been applied in face detection \citep{socolinsky2002} and latent class discovery in gene expression data \citep{priebe2003dna}.
There are several other approaches in the literature to solve the class cover problem,
including covering the classes with a set of boxes \citep{bereg2012} or set of convex hulls \citep{takigawa2009}.

In this article, we study the effects of class imbalance on two CCCD classifiers, P-CCCD and RW-CCCD.  Moreover, we report on the effects of class overlapping problem (which is defined as deterioration of classification performance when class supports overlap) along with the class imbalance problem to further investigate the performance of CCCD classifiers when imbalance and overlapping between classes co-exist.
Thus, we show that when there is a considerable amount of class imbalance, whether class supports overlap or not, the CCCD classifiers perform better than the $k$-NN classifier. We show the robustness of CCCD classifiers to the class imbalance by simulating cases having increasing levels of class imbalance. We also compare CCCD classifiers with SVM classifiers which are potentially robust to moderate levels of class imbalance but not to high levels. With respect to class imbalance problem, the $k$-NN, SVM and decision tree classifiers may be referred to as  ``weak" classifiers; that is, these methods perform weakly when there is imbalance in the data set.
However, such classifiers can be modified to address the unequal priors in a data set, and hence, can be converted to ``strong" classifiers which are potentially robust to the class imbalance problem. We show that CCCD classifiers are also inherently robust
(i.e., robust to class imbalance without any modification), and we compare the CCCD classifiers against the state-of-the-art strong classification methods which are constructed to perform well when class imbalance occurs. We consider ensemble learning, cost-sensitive learning and resampling schemes in conjunction with $k$-NN, SVM and decision tree classifiers, and show that RW-CCCDs and P-CCCDs perform comparable to these strong classifiers.

Among the two variations of CCCD classifiers, we show that the RW-CCCD is more appealing in many aspects.
For both simulated and real life examples, RW-CCCDs perform better than P-CCCDs and weak classifiers,
and perform comparable to strong classifiers when the classes of data sets are imbalanced and/or overlapping.
Moreover, we report on the complexity of the two CCCD classifiers and
demonstrate that RW-CCCDs reduce the data sets substantially more than the other classifiers,
thus increasing the testing speed.
But most importantly, while reducing the majority class to mitigate the effects of class imbalances, CCCDs preserve the information on the discarded points of the majority class.
CCCDs provide a novel potential solution to the class imbalance problem;
that is, they capture the probability density around prototype points (i.e., members of the dominating sets) as radii of the covering balls. Hence, CCCDs preserve the information while reducing the data set
In the literature, only the strong classifiers based on hybrids of ensembles and resampling schemes achieve a similar task which requires multiple classifiers to be employed, and thus, result in lengthy training and testing time. However, CCCDs define single classifiers that undersample the data set with, possibly, a slight loss of information.

We provide a short review of the existing methods for classifying data sets with class imbalance in Section~\ref{CIreview}, introduce P-CCCD and RW-CCCD classifiers in Section~\ref{cccd}, discuss the balancing effect of CCCD classifiers in Section~\ref{Balcccd}.
Finally, in Section~\ref{comparecccd},
we compare the CCCD classifiers with the classifiers that are both sensitive (weak) and non-sensitive (strong) classifiers to class imbalance by simulated and real data sets, and report on the computational complexity of all weak classifiers.

\section{Methods for Handling Class Imbalance Problem} \label{CIreview}

Solving the class imbalance problem received considerable attention in the machine learning literature \citep[see][]{chawla2004,kotsiantis2006,longadge2013}. Almost all algorithms designed to mitigate the effects of class imbalance incorporate a ``weak" classifier which is modified to show some level of robustness to the class imbalance problem. The weak algorithm is modified either (i) in data level which involves a pre-processing of the data set being used in training, or (ii) in algorithmic level such that a ``strong" classifier is constructed with a decision rule suited for the imbalances in the data set. Many modern algorithms are hybrids of both types; but in particular, there are mainly three of them: resampling methods, cost-sensitive methods, and ensemble methods \citep{he2009}.

Resampling methods are commonly employed to remove the effects of class imbalance in the classification process. Resampling methods provide solutions to the class imbalance problem by (i) downsizing the majority class (undersampling) or (ii) generating new (synthetic) points for the minority class (oversampling). Hence, such methods modify the classifiers only at the data level. It might be useful to clean or erase some points in the majority class to balance the data \citep{drummond2003,liu2009}.
However, in some cases, all points from both classes may be valuable/important, and hence, should be kept despite the differences in the class sizes. Oversampling methods generate synthetic points similar to the minority class to mitigate the class imbalance problem while preserving the information \citep{han2005}. On the other hand, \citet{batista2004} suggest that the combination of both over and undersampling methods can further improve the classification performance. One such method is the SMOTE+ENN method where the oversampling method SMOTE of \citet{chawla2002} and edited nearest neighbors (ENN) method of \citet{wilson1972} are applied to an imbalanced data set, consecutively. While SMOTE balances the classes of the data set by generating artificial points between members of the minority class, ENN cleans the data set to further increase the classification performance of the weak classifier. Here, ENN method is an undersampling method that primarily aims to remove noisy points from the data set but not to balance the classes.

Another family of methods, namely cost-sensitive learning methods, has originated from real life: the cost of misclassifying a minority and a majority class member is usually not the same \citep{elkan2001}. Frequently, the minority class has higher misclassification cost than the majority class. Classification methods such as decision trees (e.g., C4.5), can be modified to take these costs into account \citep[see][]{ling2004,zadrozny2003}. C5.0 is an extended version of C4.5 incorporating the cost of each class \citep{kuhn2013}. Most weak classifiers can be easily modified so as to recognizing misclassification costs. The constrained violation cost $C$ of SVM classifiers can be adjusted to individual class costs \citep{chang2011}. As for $k$-NN, one solution is to appoint weights to all points of the data set with respect to their classes. Hence, such weights are the costs of classes giving precedence to minority class points \citep{barandela2003}.
On the other hand, for those algorithms that costs are not inherently recognizable or available, meta-learning schemes can be used along with weak classifiers without modifying the classifiers.
Such learning methods are similar to ensemble learning methods \citep{domingos1999}.

A fast developing field called ensemble learning also contributes to the family of methods handling the class imbalance problem \citep[see][]{galar2012}. The idea is to combine several classifiers to create a new classifier which has significantly better performance than its constituents \citep{rokach2010}. AdaBoost is a popular algorithm among this family of learning methods \citep{freund1997,wu2008}. AdaBoost assigns weights to each of the points in the data set and updates these weights in accordance with how well the classes of points are estimated by each classifier. \citet{galar2012} provide a survey of the most important ensemble learning methods that solve the class imbalance problem. However, it has been observed in some studies that ensemble learning methods work best when used together with resampling methods \citep{lopez2013}.
In fact, ensembling and resampling schemes compensate the shortcomings of each other. The EasyEnsemble is a classifier with two levels of ensembles. First, a random undersampled majority class and the original minority class are used to train an ensemble classifier, then another random sample is drawn in the same way to train a second ensemble. This process is repeated several times to mitigate the effects of information loss as each ensemble would be applied on a different random subset of the majority class.

\section{Classification with Class Cover Catch Digraphs}
\label{cccd}

Class Cover Catch Digraphs (CCCDs) offer graph theoretic solutions to CCP \citep{priebe2001,priebe2003}.
The objective of CCP is to find a region that covers the members of a specific class. More specifically, let $(\Omega,M)$ be a measurable space and let $\mathcal{X}_n=\{x_1,x_2,...,x_n\} \subset \Omega$ and $\mathcal{Y}_m=\{y_1,y_2,...,y_m\} \subset \Omega$ be observations from two classes $\mathcal{X}$ and $\mathcal{Y}$ with class conditional distributions $F_X$, $F_Y$ and a joint cdf $F_{X,Y}$, respectively. Let $\Omega=\mathbb{R}^d$ and, without loss of generality, assume that the target class (i.e., the class of interest) is $\mathcal{X}$.
In a CCCD, for $x_i,x_j \in \mathcal{X}_n \subset \mathbb{R}^d$, $x_i$ is the center of a ball with radius $r_i=r(x_i)$
which is a function of $d(x_i,y_j)$ for $y_j \in \mathcal{Y}_m$.
Each ball is represented by $B_i=B(x_i,r_i)$ and if $x_j \in B_i$ then $x_i$ is said to cover (or catch) $x_j$.
Here, $d(.,.)$ can be any dissimilarity measure but we use the Euclidean distance henceforth.
A CCCD is a digraph $D=(V,A)$ with vertex set $V=\mathcal{X}_n$ and the arc set $A$ where $(x_i,x_j) \in A$ if and only if $x_j \in B_i$.
The term ``catch" refers to arc $(x_i,x_j)$ of the digraph $D$ where $x_i$ is said to catch $x_j$.
The binary relation $x_i \sim x_j$, which is defined as $x_j \in B_i$, is asymmetric,
thus the adjacency of $x_i$ and $x_j$ is represented with directed edges or arcs which yield a  digraph instead of a graph.

In CCCDs, the goal is to find a subset of balls $\mathcal C_{\mathcal{X}} \subseteq \mathcal B_{\mathcal{X}}=\{B_1,B_2,...,B_n\}$ such that $Q_{\mathcal{X}} \subseteq \cup_{B \in \mathcal C_{\mathcal{X}}} B$ for $Q_{\mathcal{X}} \subseteq \mathcal{X}_n$
and $C_{\mathcal{X}} = \cup_{B \in \mathcal C_{\mathcal{X}}}$ is the cover of $Q_{\mathcal{X}}$.
Here the set $Q_{\mathcal{X}}$ is some desirable subset of the target class training set $\mathcal{X}_n$ which we want to cover.
Preferably, the goal is to find a set $\mathcal C_{\mathcal{X}}$ such that $Q_{\mathcal{X}} = \mathcal{X}_n$, however it might be desirable that the class cover may ignore some target class points to avoid overfitting. If a class cover of a CCCD fails to cover some target class points, it is called an \emph{improper} cover, otherwise it is a \emph{proper} cover.
For covering $\mathcal{Y}_m$, we reverse the roles of classes $\mathcal{X}$ and $\mathcal{Y}$;
the class $\mathcal{Y}$ becomes the target class and $\mathcal{X}$ becomes the non-target class. Finding an appropriate cover $C_{\mathcal{X}}$ is equivalent to finding the dominating set of the CCCD with $V=\mathcal{X}_n$.
Let $N(s) = \{t \in V: (s,t) \in A\}$ be the \emph{open neighborhood} of a vertex $s \in V$:
the set of vertices to which there is an arc from the vertex $s$, or the neighbors of $s$. A dominating set of a digraph $D$ is defined as a subset of vertices $S \subseteq V$ such that union of the closed neighborhoods, defined by $\bar{N}(s)=N(s) \cup \{s\}$, of elements of $S$ is the vertex set of the digraph: $\cup_{s \in S}\bar{N}(s) = V$.
Among all dominating sets, usually the ones with minimum cardinality, called the \emph{minimum dominating sets}, are preferable.
The cardinality of the minimum dominating set(s) is referred to as the \emph{domination number}, denoted as $\gamma(D)$. However, minimum dominating sets are often computationally intractable and finding them is, in general, an NP-hard optimization problem. Hence, greedy algorithms are often employed to find sets with approximately minimum cardinality \citep{chvatal1979, devinney2003}.

CCCDs can easily be generalized to the multi-class case with $k$ classes.
We transform the multi-class case into a two-class setting and find the cover of $i$-th class, $C_i$, for each $i=1,\cdots,k$.
To establish the set of covers $\mathcal{C} = \{C_1,C_2, \cdots, C_k\}$
associated with a set of classes $\mathfrak{X} = \{\mathcal{X}_1,\mathcal{X}_2,\cdots,\mathcal{X}_k\}$,
for class $i$,
we merge the classes into two classes as $\mathcal{X}_T=\mathcal{X}_i$ and $\mathcal{X}_{NT}=\cup_{i \neq j} \mathcal{X}_j$ for $i,j=1,\cdots,k$.
We take the classes $\mathcal{X}_{T}$ and $\mathcal{X}_{NT}$ as target class and non-target class, respectively.
More specifically, as before, target class is the class we want to find the cover of,
and the non-target class is the union of the remaining classes.

We employ two families of CCCDs, pure-CCCDs (P-CCCDs) and random walk CCCDs (RW-CCCDs) that differ in the definition of the radius $r(x)$.
In these two digraphs,
the (approximate) minimum dominating set $S$ and the classifier are defined in slightly different ways;
with the main distinction between the two being the way the covers are defined.
The covering balls of P-CCCDs do not contain any non-target class point (hence the name ``pure") whereas RW-CCCDs possibly allow some non-target class points inside of the class cover of the target class so as to avoid overfitting.
Moreover, some target class points may also be missed by the covers of RW-CCCDs. Therefore, P-CCCDs construct pure and proper covers but RW-CCCD covers are not necessarily pure or proper.

\subsection{Classification with P-CCCDs}

In P-CCCDs, the covering balls $B_x=B(x,r(x))$ exclude all non-target class points.
Thus, for a target class point $x \in \mathcal{X}_n$,
which is the center of a ball $B_x$, the radius $r(x)$
should be smaller than or equal to the distance between $x$ and the closest non-target point
$y \in \mathcal{Y}_m$: $r(x) \le  \min_{y \in \mathcal{Y}_m} d(x,y)$. Given $\tau \in (0,1]$, the radius $r(x)$ is defined as follows \citep{marchette2010}:
\begin{equation} \label{equtau}
	r(x):=(1-\tau)d(x,l(x)) + \tau d(x,u(x)),
\end{equation}
where
\begin{equation*}
	u(x):=\argmin_{y \in \mathcal{Y}_m} d(x,y)
\end{equation*}
and
\begin{equation*}
	l(x):=\argmax_{z \in \mathcal{X}_n} \{d(x,z): d(x,z) < d(x,u(x))\}.
\end{equation*}

\begin{figure}
\centering
\begin{tabular}{cc}
\includegraphics[scale=0.23]{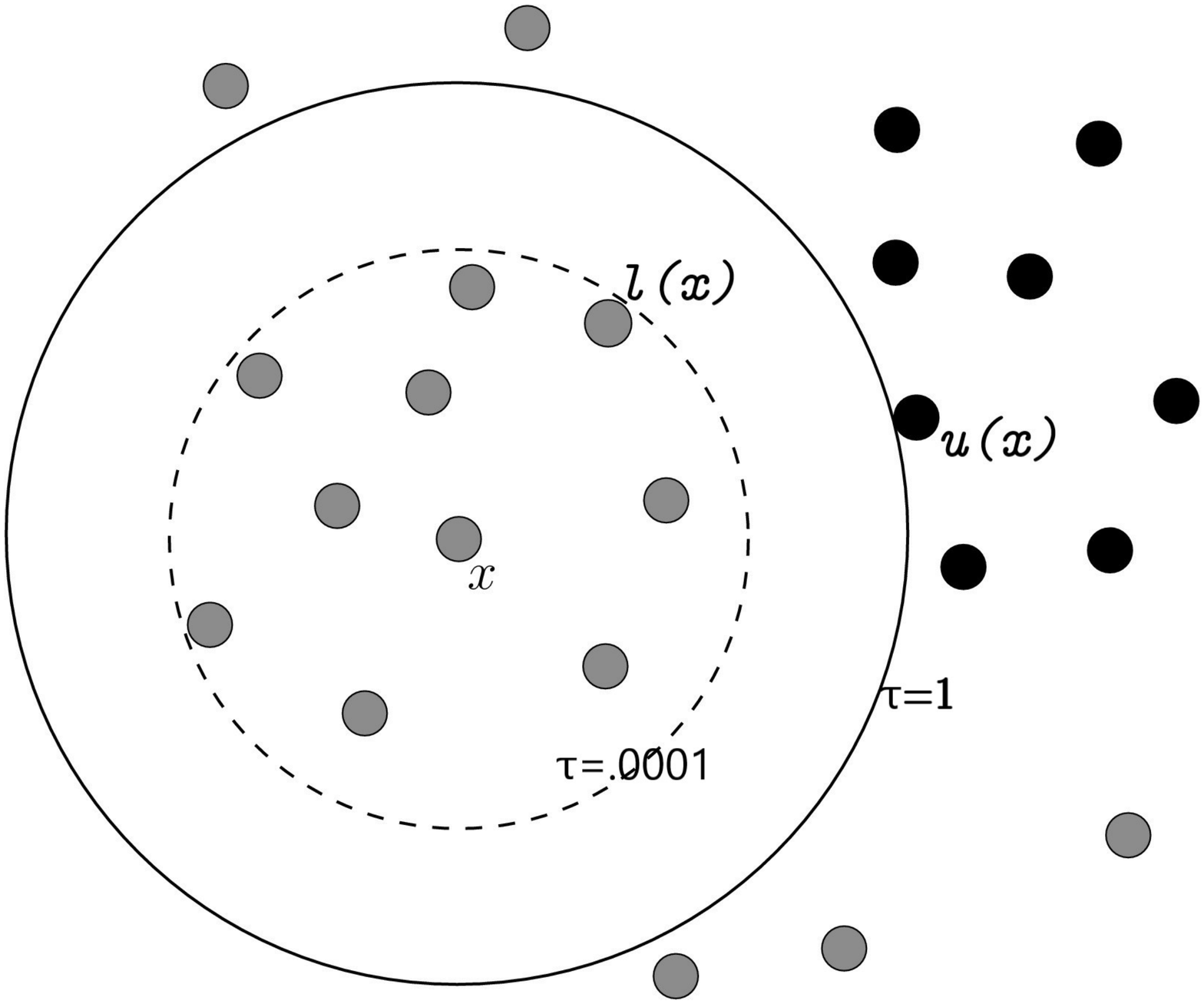}
\end{tabular}
\caption{The effect of $\tau$ on the radius $r(x)$ of the target class point $x$ in a two-class setting.
Grey and black points represent the target and non-target classes, respectively.
The solid circle centered at $x$ is constructed with the radius associated with
$\tau=1$ and the dashed one with $\tau=0.0001$ \citep{devinney2003}.}
\label{tau}
\end{figure}

\begin{figure}
\centering
\begin{tabular}{cc}
\includegraphics[scale=0.40]{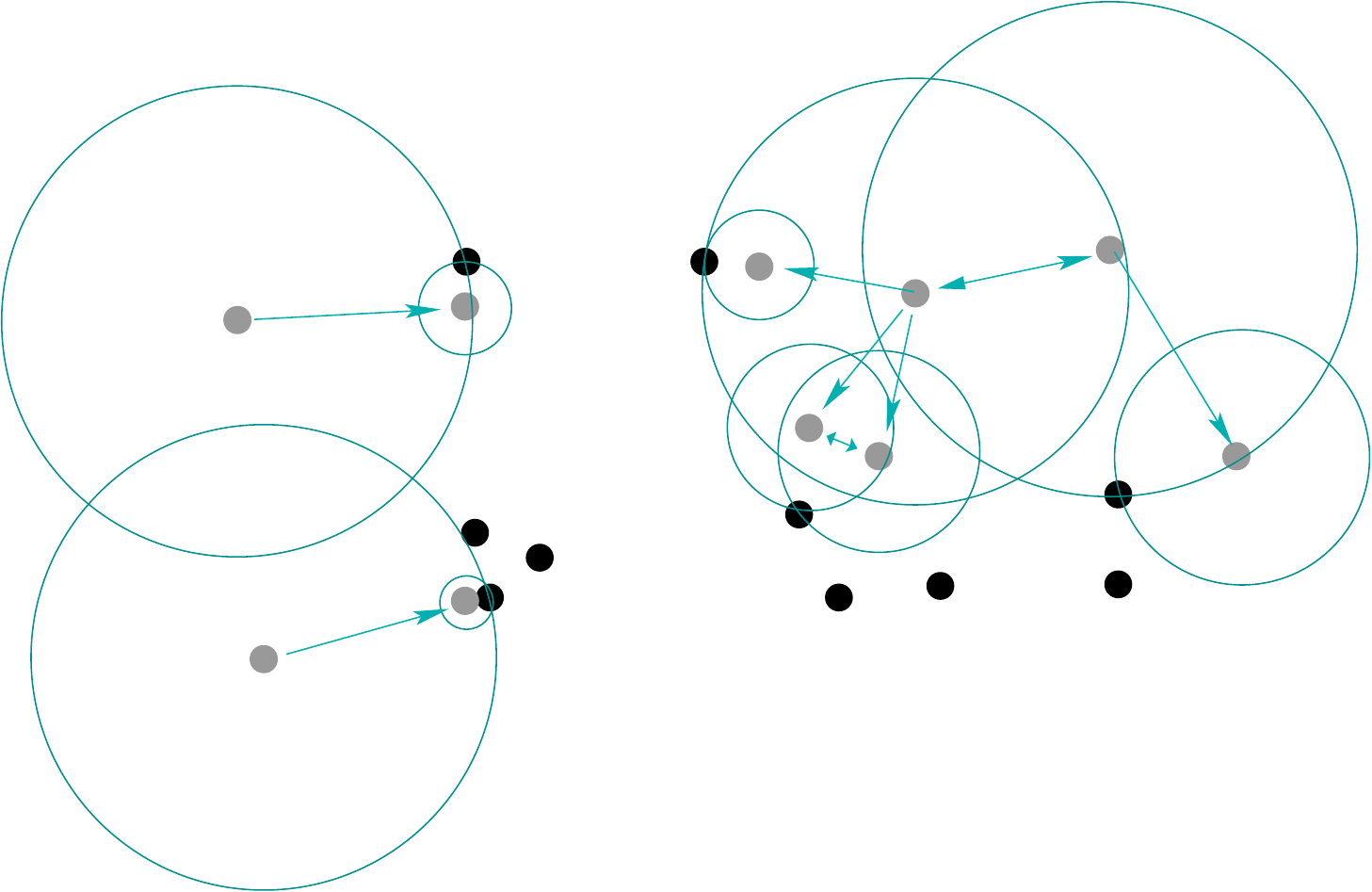} & \includegraphics[scale=0.40]{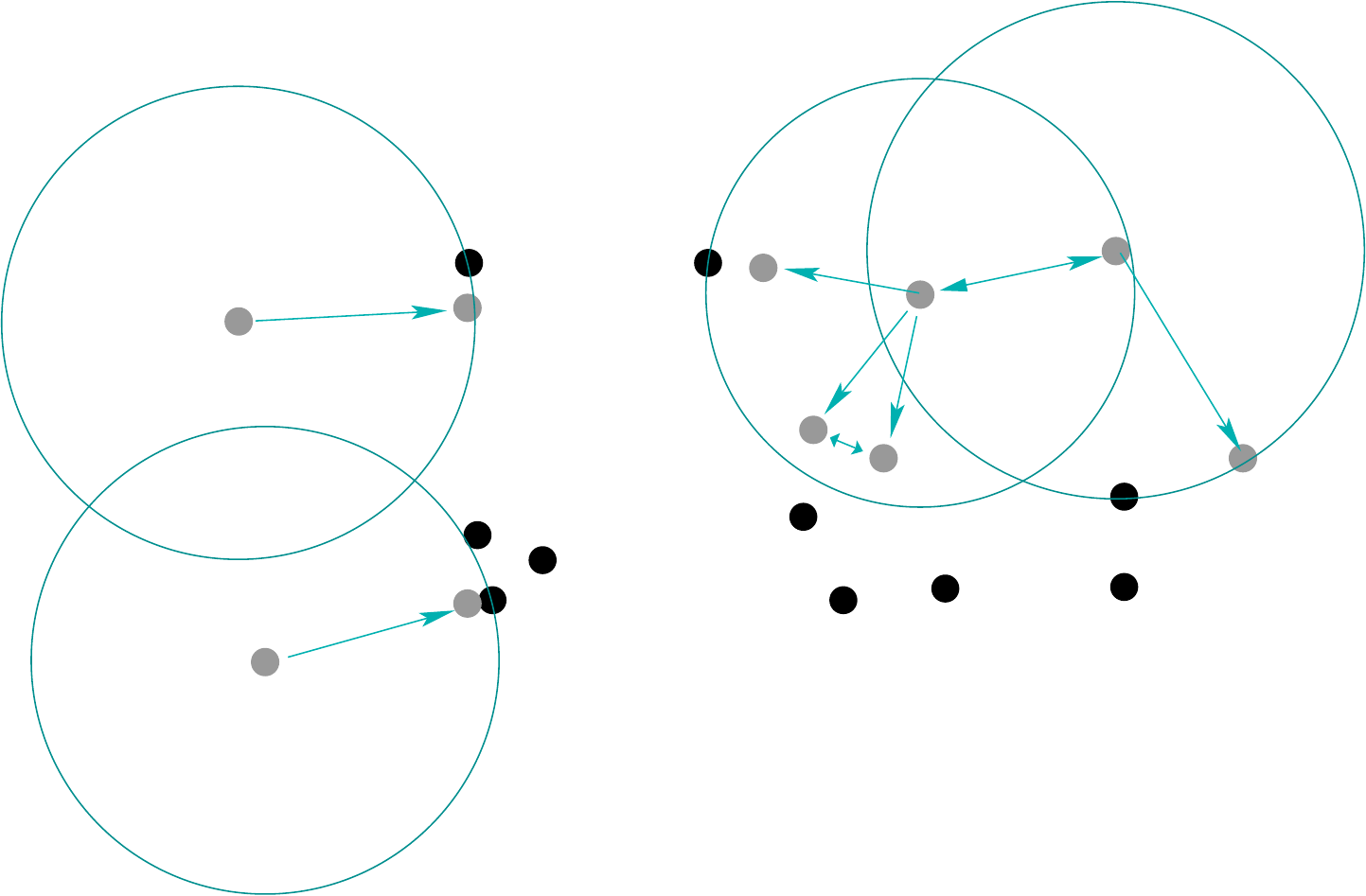}\\
\\
\multicolumn{2}{c}{\includegraphics[scale=0.40]{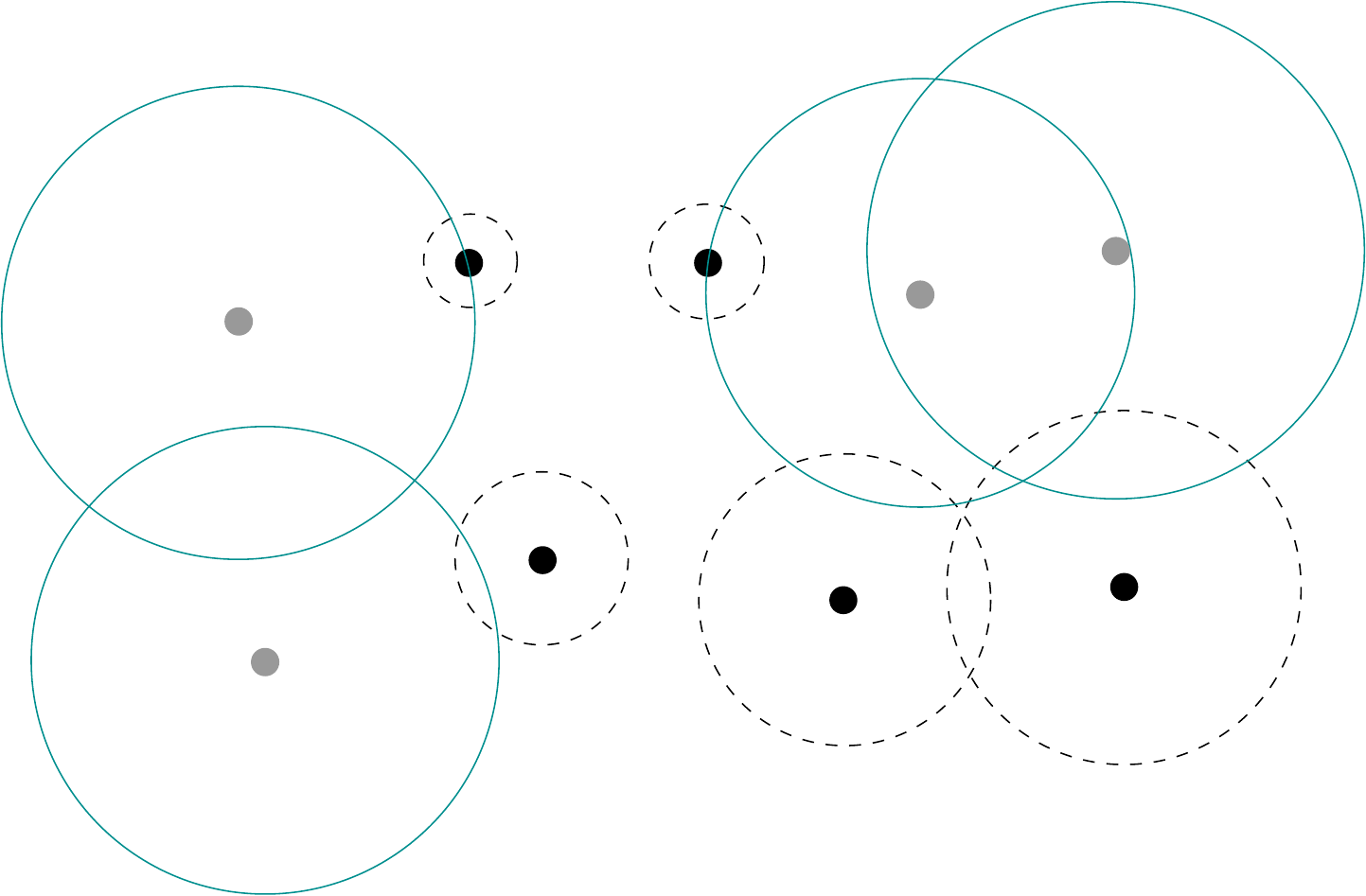}} \\
\end{tabular}
\caption{An illustration of the CCCDs (with the grey points representing the points from the target class) in a two-class setting.
Presented in top left are all covering balls and the digraph $D=(V,A)$ and
in top right are the balls that constitute a minimum cardinality class cover for the target class and
are centered at points which are the elements of the minimum dominating set $S \subseteq V$.
In the bottom panel,
we present the dominating sets of both classes and their associated balls which establish the class covers.
The class cover of grey points is the union of solid circles,
and that of black points is the union of dashed circles.}
\label{fig}
\end{figure}

The effect of parameter $\tau$ on the radius $r(x)$ is illustrated in Figure~\ref{tau} \citep{devinney2003}.
The ball with radius $r(x)$ catches the neighboring target class points,
and for any $\tau \in (0,1]$, the ball $B_x$ catches the same points as well.
Hence, the choice of $\tau$ does not affect the structure of digraph but might affect
the classification performance which will be shown later in Section~\ref{comparecccd}. On the other hand, for all $x \in \mathcal{X}_n$, the definition of $r(x)$ in Equation~(\ref{equtau}) keeps any non-target point $y \in \mathcal{Y}_m$ out of the ball $B_x$,
that is $ B_x \cap \mathcal{Y}_m = \emptyset$ for all $B_x \in \mathcal C_{\mathcal{X}}$.
Here, $B_x$ is an open ball: $B_x = \{z \in \mathbb{R}^d: d(x,z) < r(x)\}$.
The digraph $D$ is ``pure" since the balls contain only the target class points;
hence, the name pure-CCCD.
Once all balls are constructed, so is the digraph $D$.
Therefore, we have to find the set of covering balls $\mathcal C_\mathcal{X}$ which is equivalent to finding a minimum dominating set $S \subseteq V$.
The greedy algorithm of finding an approximate minimum dominating set of a P-CCCD is given in Algorithm~\ref{greedy}.
At each iteration, the vertex which has the largest neighborhood (i.e., highest number of outgoing arcs)
is removed from the graph together with its neighbors.
Then, the process is repeated until all vertices of $D$ are removed. The algorithm adds elements to the dominating set until all points are either dominated or dominate some other points. Hence, the covers established by P-CCCDs are proper covers: $Q_{\mathcal{X}}=\mathcal{X}_n$ and $Q_{\mathcal{Y}}=\mathcal{Y}_m$.
The P-CCCD of one class, its associated class cover (constructed by the elements of the dominating set), and covers of both classes are illustrated in Figure~\ref{fig}.

\begin{algorithm}[h]
\small
\begin{algorithmic}[1]
 \REQUIRE A digraph $D=(V,A)$
 \ENSURE An approximate minimum dominating set, $S$
 \STATE  \textbf{set} $H=V$ and $S=\emptyset$
 \WHILE{$H \neq \emptyset$}
  \STATE $v^{*} = \argmax_{v \in V} |\bar{N}(v)|$
  \STATE $S = S \cup \{v^{*}\}$
  \STATE $H = V \setminus \bar{N}(v^{*})$
  \STATE $D=D[H]$
 \ENDWHILE
\end{algorithmic}
\caption{The greedy algorithm for finding an approximate minimum dominating set of a digraph $D$.
Here, $D[H]$ is the digraph induced by the set of vertices $H \subseteq V$ \citep[see][]{west2000}.}
\label{greedy}
\end{algorithm}

Before Algorithm~\ref{greedy} finds an approximate solution, we should first construct the digraph $D$. The P-CCCD cover $C_{\mathcal{X}}$ and the P-CCCD $D$ depend on the distances between points of the target class $\mathcal{X}_n$, denoted by the matrix $\mathcal{M}_{\mathcal{X}}$, and the distances from all points of $\mathcal{X}_n$ to all points of $\mathcal{Y}_m$, denoted by matrix $\mathcal{M}_{\mathcal{X},\mathcal{Y}}$. Later, we construct the set of balls $B_{\mathcal{X}}=\{B_1,B_2,...,B_n\}$, and get the set of arcs $A$ where $V=\mathcal{X}_n$. Hence, the minimum cardinality ball cover problem is reduced to a minimum dominating set problem.
We find such a cover with Algorithm~\ref{p_greedy} which runs in quadratic time and, in addition, depends on the dimensionality of the training set $\mathcal{X}_n \cup \mathcal{Y}_m$.
	
\begin{algorithm}[h]
\small
\begin{algorithmic}[1]
 \REQUIRE Points of the target class $\mathcal{X}_n$, the non-target class $\mathcal{Y}_m$ and the P-CCCD parameter $\tau \in (0,1]$
 \ENSURE An approximate minimum cardinality ball cover $C_{\mathcal{X}}$
 \STATE $r(x):=(1-\tau)d(x,l(x)) + \tau d(x,u(x))$ for all $x \in \mathcal{X}_n$	
 \STATE Construct the digraph $D$ with the set $B_{\mathcal{X}}$.
 \STATE Find the approximate minimum dominating set $S$ of digraph $D$ by Algorithm~\ref{greedy}.
 \STATE $C_{\mathcal{X}}:=\cup_{s \in S} B(s,r(s))$
\end{algorithmic}
\caption{The greedy algorithm for finding an approximate minimum cardinality ball cover $C_{\mathcal{X}}$ of the
target class points $\mathcal{X}_n$ given a set of non-target class points $\mathcal{Y}_m$.}
\label{p_greedy}
\end{algorithm}
	
\noindent
{\bf Theorem 1} {\it Algorithm~\ref{p_greedy} is an $\mathcal{O}(\log n)$-approximation algorithm and finds an approximate minimum cardinality ball cover $C_{\mathcal{X}}$ of the target class $\mathcal{X}$ in $\mathcal{O}(n(n+m)d)$ time.}

\noindent
{\bf Proof}. The algorithm is polynomial time reducible to a greedy minimum set cover algorithm which finds an approximate solution with size at most $\mathcal{O}(\log n)$ times of the optimum solution \citep{chvatal1979,cannon2004}.
We first calculate the distance matrices $\mathcal{M}_{\mathcal{X}}$ and $\mathcal{M}_{\mathcal{X},\mathcal{Y}}$ which take $\mathcal{O}(n^2 d)$ and $\mathcal{O}(nmd)$ time, respectively.
Constructing the digraph $D$ requires computing $l(x)$ and $u(x)$ in Equation~(\ref{equtau}) for all $x \in \mathcal{X}_n$,
taking $\mathcal{O}((n^2+nm)d)$ time in total.
Then, we set the arc set $A$ in $\mathcal{O}(n^2)$ time.
Finally, the algorithm finds a solution for the digraph $D$ in $\mathcal{O}(n^2)$ time, hence the total running time of the algorithm is $\mathcal{O}(n(n+m)d)$. \hfill\ensuremath{\square}
	
 When $\mathcal{Y}$ is the target class, observe that the time complexity is $\mathcal{O}(m(n+m)d)$, and an approximate solution is of size at most $\mathcal{O}(\log{m})$ times the optimal solution by Theorem 1, since $m=|\mathcal{Y}_m|$. A P-CCCD classifier consists of the covers of all classes, hence the total training time of finding covers based on CCCDs of a data set in a two-class setting is $\mathcal{O}((n+m)^2d)$.

\indent After establishing both class covers $C_{\mathcal{X}}$ and $C_{\mathcal{Y}}$, any new data point can be classified in $\mathbb{R}^d$ according to where it resides. Here, there are three cases according to the location of the given point, $z$, to be classified: $z$ is (i) only in $C_{\mathcal{X}}$ or $C_{\mathcal{Y}}$, (ii) in both $C_{\mathcal{X}}$ and $C_{\mathcal{Y}}$ or (iii) in neither of $C_{\mathcal{X}}$ and $C_{\mathcal{Y}}$. The case (i) is straightforward: $z$ belongs to class $\mathcal{X}$ if  $ z \in C_{\mathcal{X}} \setminus C_{\mathcal{Y}}$ or to class $\mathcal{Y}$ if $ z \in C_{\mathcal{Y}} \setminus C_{\mathcal{X}}$.
For cases (ii) and (iii),
we need to find a way to determine the class of the point in a reasonable way.
In fact, for all the cases, the estimated class of a given point $z$ is determined by
\begin{equation} \label{func}
	\argmin_{U \in \{\mathcal{X},\mathcal{Y}\}} \left[ \min_{x:B(x,r) \in \mathcal C_U} \rho(z,x) \right]
\end{equation}
	
\noindent where $\rho(z,x)=d(z,x)/r(x)$ \citep{marchette2010}.
The comparison of the dissimilarity measure $\rho(z,x)$ with 1
indicates whether or not the point $z$ is in the ball of radius $r(x)$ with center $x$,
since $\rho(z,x) \leq 1$ if $z$ is inside the (closure of the) ball and $>1$ otherwise.
The measure $\rho: \Omega \times \Omega \rightarrow \mathbb{R}_{+}$ is simply a scaled dissimilarity measure, since Euclidean distance between two points, $d(x,y)$, is divided (or scaled) with the radius, $r(x)$ or $r(y)$.
This measure violates the symmetry axiom among metric axioms since $\rho(x,y) \neq \rho(y,x)$ whenever $r(x) \neq r(y)$.  However, \citet{priebe2003} showed that the dissimilarity measure $\rho$ satisfies the continuity condition, i.e.,
under the assumptions that both $F_X$ and $F_Y$ are continuous and strictly separable
(i.e., $\inf_{x \in s(F_X),y \in s(F_Y)} d(x,y) = \delta > 0$
where $s(F_X)$ and $s(F_Y)$ are the supports of the classes $\mathcal{X}$ and $\mathcal{Y}$, respectively), P-CCCD classifiers are consistent; that is, their misclassification error approaches to the Bayes optimal classification error as $m,n \rightarrow \infty$.
The measure $\rho$ favors points with bigger radii;
that is, for example, for a new point $z$ equidistant to two points, the point with bigger radius is closer in terms of this scaled dissimilarity measure;
for example, $\rho(z,x) < \rho(z,y)$ when $d(x,z)=d(y,z)$ and $r(x) > r(y)$. The radius $r(x)$ can be viewed as an indicator of the density around the point $x$.
Thus, a point $x$ with bigger radius might suggest that the point $z$ is
more likely be drawn from the same distribution (or class) where $x$ is drawn (i.e., from the class denser in probability).

\subsection{Classification with Random Walk CCCDs}

\begin{figure}
\centering
\begin{tabular}{|c|}
\hline
\includegraphics[scale=0.14]{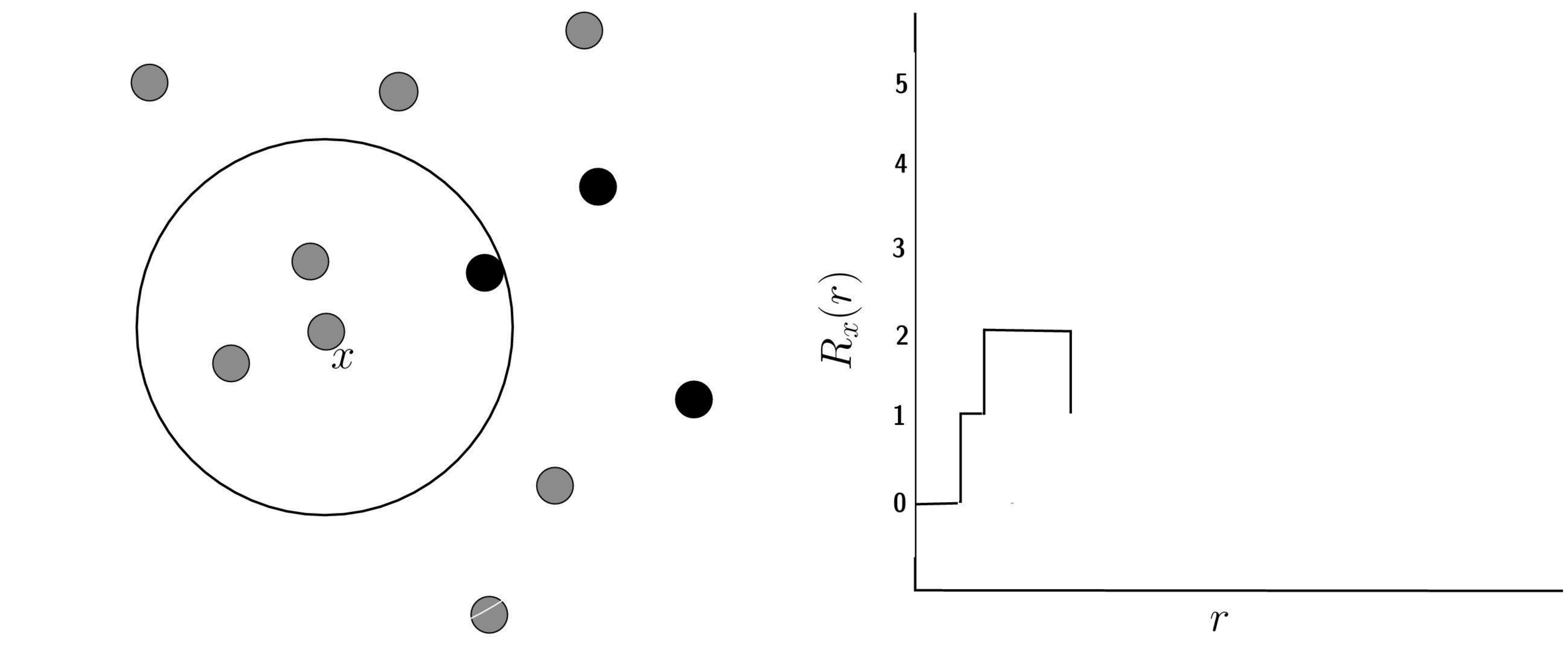} \\
\hline
\includegraphics[scale=0.14]{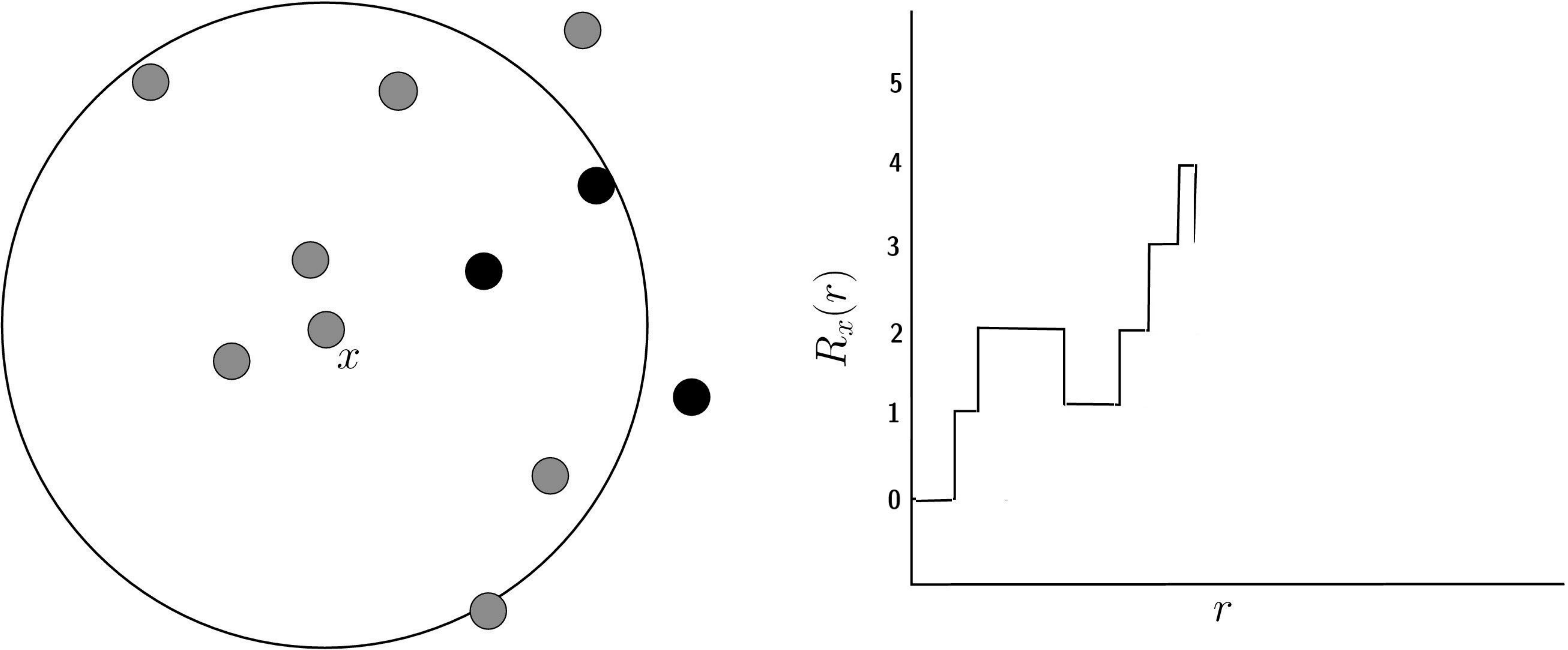} \\
\hline
\end{tabular}
\caption{Two snapshots of $R_{x}(r)$ associated with the ball $B_{x}$ centered at $x$ for $m=n$;
only a subset of the study region is shown.}
\label{randomwalk}
\end{figure}

For P-CCCDs, the class covers defined by CCCDs are ``pure" of non-target class points;
that is, no member of the non-target class is allowed inside the cover of the target class.
As in Figure~\ref{tau}, the ball centered at the point $x$ cannot expand any further
since its radius is restricted by the distance to the closest non-target class point.
This strategy may cause the cover to overfit or be sensitive to noise or outliers in the non-target class.
By allowing some neighboring non-target class points inside the cover and some target class points outside the cover,
the random walk CCCDs (RW-CCCDs) catch as much target class points as possible
with an adaptive strategy of choosing the radii \citep{devinney2002}.
For $x \in \mathcal{X}_n$, $|\mathcal{X}_n|=n$ and $|\mathcal{Y}_m|=m$, RW-CCCDs define a function on radius of a ball given by
\begin{align}
\begin{split}
	R_{x}(r) &= R_{x}(r;\mathcal{X}_n,\mathcal{Y}_m) \\
	         &:= \frac{m}{n}|\{z \in \mathcal{X}_n: d(x,z) \leq r\}| - |\{z \in \mathcal{Y}_m: d(x,z) \leq r\}|. \\
\end{split}
\end{align}
\noindent
where second and third arguments in $R_{x}(r;\mathcal{X}_n,\mathcal{Y}_m)$ are suppressed when there is no ambiguity.
The function $R_{x}(r)$ can be viewed as a one-dimensional random walk. When the ball centered at $x \in \mathbb{R}^d$ expands, it hits either a target class point or a non-target class point which increases or decreases the random walk by one unit, respectively. The ratio $m/n$ is included in the first term as to avoid the bias resulted by unequal sample sizes (i.e., class imbalance). An illustration is given in Figure~\ref{randomwalk} for the case $m=n$. The function $R_{x}(r)$ aims to find such radii that it contains a few non-target class points and sufficiently many target class points. In addition, we also want to avoid balls with large radii. Hence, the radius of $x$ is the value maximizing $R_{x}(r)$ with an additional penalty function $P_{x}(r)$ which biases toward small radii:
\begin{equation} \label{rwequ}
 r_{x} := \argmax_{r \in \{d(x,z): z \in \mathcal{X}_n \cup \mathcal{Y}_m\}} R_{x}(r) - P_{x}(r).
\end{equation}

\noindent
Although a penalty function seems fit, \citet{devinney2003} pointed out that
the choice of $P_{x}(r)=0$ usually works sufficiently well in practice.
As in P-CCCDs, the radius of a ball represents the (probability) density of its center's neighborhood.
Maximizing $R_{x}(r)$ determines the best possible radius.
Moreover, unlike P-CCCDs, the balls of RW-CCCDs are closed balls: $\bar B_x = \{z \in \mathbb{R}^d: d(x,z) \leq r(x)\}$.

Similar to P-CCCDs, finding a cover, or a dominating set, of a RW-CCCD is an NP-hard problem.
However, RW-CCCDs find the minimum dominating sets in a slightly different fashion. Instead of finding a set $S$ such that $\cup_{s \in S}\bar{N}(s) = V$ as in Algorithm~\ref{greedy}, we first locate the vertex $x^*$ (a target class point) which has maximum of some score, $T_{x^*}$, and remove all target and non-target class points covered with the ball of this vertex, $\bar B_{x^*}=\bar B(x^*,r^*)$.
In the next iteration, we recalculate the radii of remaining target class points, find the next point with the maximum score and continue until all target class points are covered. This greedy method of finding dominating set(s) $S$ of RW-CCCDs is given in Algorithm~\ref{rw_greedy}. The resulting dominating set $S$ has approximate minimum cardinality. For each target class point $x \in \mathcal{X}_n$, the score $T_{x}$ is associated with $R_{x}(r_{x})$ and is given by
\begin{equation} \label{rwscore}
	T_{x} = R_{x}(r_{x}) - \frac{r_{x} n_u}{2 d_{m}(x)}
\end{equation}

\noindent where $n_u$ is the number of uncovered target class points in the current iteration,
and $d_{m}(x)=\max_{z \in \mathcal{X}_n} d(x,z)$.
The term which is linear in $r_{x}$ of the right hand side of Equation~(\ref{rwscore}) is similar to $P_x(r)$ in Equation~(\ref{rwequ}): it biases the scores toward choosing dominating points with smaller radii.
On the other hand, Algorithm~\ref{rw_greedy} is likely to choose dominating points with radius $r=0$. These points only dominate themselves but they are thought of being not covered since their balls have radii $r=0$.
Hence, RW-CCCDs may establish improper covers.

\begin{algorithm}[h]
\small
\begin{algorithmic}[1]
 \REQUIRE Target class points $\mathcal{X}_n$ and non-target class points $\mathcal{Y}_m$
 \ENSURE Approximate minimum dominating set $S$ of $\mathcal{X}_n$
 \STATE  $H_0=\mathcal{X}_n$, $H_{1}=\mathcal{Y}_m$ and $S = \emptyset$
 \STATE  $\forall x \in \mathcal{X}_n$, $d_{m}(x)=\max_{z \in \mathcal{X}_n} d(x,z)$
 	\WHILE{$H_0 \neq \emptyset$}
 		\STATE $n_u=|H_0|$
 		\FORALL{$x \in H_0$}
 			\STATE $r(x)=\argmax_{r} R_x(r;H_0,H_1)$ for $r \in \{d(x,z): z \in H_0 \cup H_1\}$ 	
 		\ENDFOR  		
  		\STATE $x^* = \argmax_{x \in H_0} T_x$ %R_x(r(x);H_0,H_1) - \frac{r(x) n_u}{2 d_m(x)}$
  		\STATE $S = S \cup \{x^*\}$
  		\STATE $H_0 = H_0 \setminus (\bar B_{x^*} \cap \mathcal{X}_n)$ and $H_1 = H_1 \setminus (\bar B_{x^*} \cap \mathcal{Y}_m)$
 	\ENDWHILE
 	\STATE $C_{\mathcal{X}}:=\cup_{s \in S} B(s,r(s))$
\end{algorithmic}
\caption{The greedy algorithm for finding an approximate minimum dominating set for RW-CCCDs of points $\mathcal{X}_n$
from the target class given non-target class points $\mathcal{Y}_m$.}
\label{rw_greedy}
\end{algorithm}

Algorithm~\ref{rw_greedy} is similar to Algorithm~\ref{p_greedy},
however after each iteration,
a point is added to the set $S$ and the random walk $R_x(r)$ is recalculated for all uncovered $x \in H_0$. Hence, we need an additional sweep on the training set which makes Algorithm~\ref{rw_greedy} run in cubic time.

\noindent
{\bf Theorem 2} {\it Algorithm~\ref{rw_greedy} finds (possibly improper) covers $C_{\mathcal{X}}$ of the target class $\mathcal{X}$ in $\mathcal{O}((n+d+\log{(n+m)})(n+m)^2)$ time.}

\noindent
{\bf Proof}.
In Algorithm~\ref{rw_greedy},
the matrix of distances between points of training set $\mathcal{X}_n \cup \mathcal{Y}_m$ must be computed since, for all $x \in \mathcal{X}_n \cup \mathcal{Y}_m$, the entire data set is swept to maximize $R_x(r)$. This takes $\mathcal{O}((n+m)^2d)$ time. The algorithm runs until all target class points are covered, but for each iteration, the random walk $R_x(r)$ is recalculated.
The maximum $R_x(r_x)$ could be found by sorting the distances for all $x \in H_0$ which could be done prior to the while loop.
This sorting takes $\mathcal{O}((n+m)^2 \log{(n+m)})$ time.
Since $H_0$ and $H_1$ are updated at each iteration,
we can just erase the distances corresponding to points covered by $\bar B_{x^*}$ which does not change the order of sorted list provided before the while loop. Hence, $\argmax R_x(r_x)$ is found and the covered points erased in $\mathcal{O}((n+m)^2)$ time. The while loop iterates $n$ times in the worst case, and hence the algorithm runs in a total of $\mathcal{O}((n+d+\log{(n+m)})(n+m)^2)$ time. \hfill\ensuremath{\square}

Note that Algorithm~\ref{rw_greedy} finds a cover of $\mathcal{Y}_m$ in $\mathcal{O}((m+d+\log{(n+m)})(n+m)^2)$ time which makes a RW-CCCD classifier trained in $\mathcal{O}((n+m)^3)$ time for $d < n$ and $\log{(n+m)} < n$. RW-CCCD classifiers are much better classifiers that potentially avoid overfitting, but with a cost of being much slower compared to the P-CCCD classifiers.

Since P-CCCD covers are pure and proper covers, P-CCCD classifiers tend to overfit \citep{devinney2003}. In RW-CCCDs, covering balls allow some points of $\mathcal{Y}_m$ inside $C_{\mathcal{X}}$ to increase average classification performance.
In that case, Algorithm~\ref{rw_greedy} cannot be reduced to a minimum set cover problem,
since the definition of sets change after adding a single point to the dominating set. Hence, the upper bound $\mathcal{O}(\log{n})$ does not apply to RW-CCCDs. However, we expect to get bigger balls in RW-CCCDs compared the ones in P-CCCDs which intuitively suggests that the covers of RW-CCCDs are lower in cardinality. We conduct empirical studies to show that RW-CCCDs, in fact, produce dominating sets with lower size compared to P-CCCDs in some cases.

In RW-CCCD, once the class covers (or dominating sets) are determined, the scaled dissimilarity measure in Equation~(\ref{func}) is a good choice for estimating the class of a new point $z$. However, \citet{devinney2003} incorporates the scores of each ball to produce better performing classifiers. Hence, the class of a new point $z$ is determined by
\begin{equation*}
	\argmin_{U \in \{\mathcal{X},\mathcal{Y}\}} \left[ \min_{x:\bar B(x,r) \in \mathcal C_U}  \rho(z,x)^{T_x^e} \right]
\end{equation*}

\noindent where $\rho(z,x)$ is defined as in Equation~(\ref{func}). Here, $e \in [0,1]$ controls at what level the score $T_x$ is incorporated. We observe that for $d(z,x) < r(x)$,
it follows that $\rho(z,x)^{T_x^e} = (d(z,x)/r(x))^{T_x^e}$ decreases as $T_x$ increases. Hence, if a new point $z$ is in both covers, $z \in C_{\mathcal{X}} \cap C_{\mathcal{Y}}$, the score $T_x$ is a good indicator to which class the new point $z$ belongs since the bigger the $T_x$, the more likely the ball contains more target class points. For $e=1$, we fully incorporate each score $T_x$ of covering balls and with $e=0$, we ignore the scores. By introducing a value for the parameter $e$ in $(0,1)$, it is possible to further improve the performance of RW-CCCD classifiers.

\section{Balancing the Class Sizes with CCCDs} \label{Balcccd}
	
The CCCD classifiers substantially reduce the number of majority class observations in a data set. The reason is that balls of majority class members are more likely to catch neighboring points of the same class. The greedy algorithm given in Algorithm~\ref{greedy} selects vertices with the largest closed neighborhood.
Similarly, Algorithm~\ref{rw_greedy} selects vertices so that their balls are as dense as possible (i.e., target class points are abundant in the balls)
with some contaminating non-target class points.
Both algorithms choose balls with a large number of target class points, and hence substantially reduce the data set (in particular, majority class points). Points of the minimum dominating set correspond to the centers of balls that establish the class covers.
Hence CCCD classifiers can also be viewed as \emph{prototype selection} methods
where the objective is finding a set of points, or \emph{prototypes}, $S$; from the training set to preserve or increase the classification performance while substantially reducing the sample size. However, the radii of dominating set(s) are also stored and used in the classification process.
	
\begin{figure}
\centering
\begin{tabular}{cc}
\includegraphics[scale=0.40]{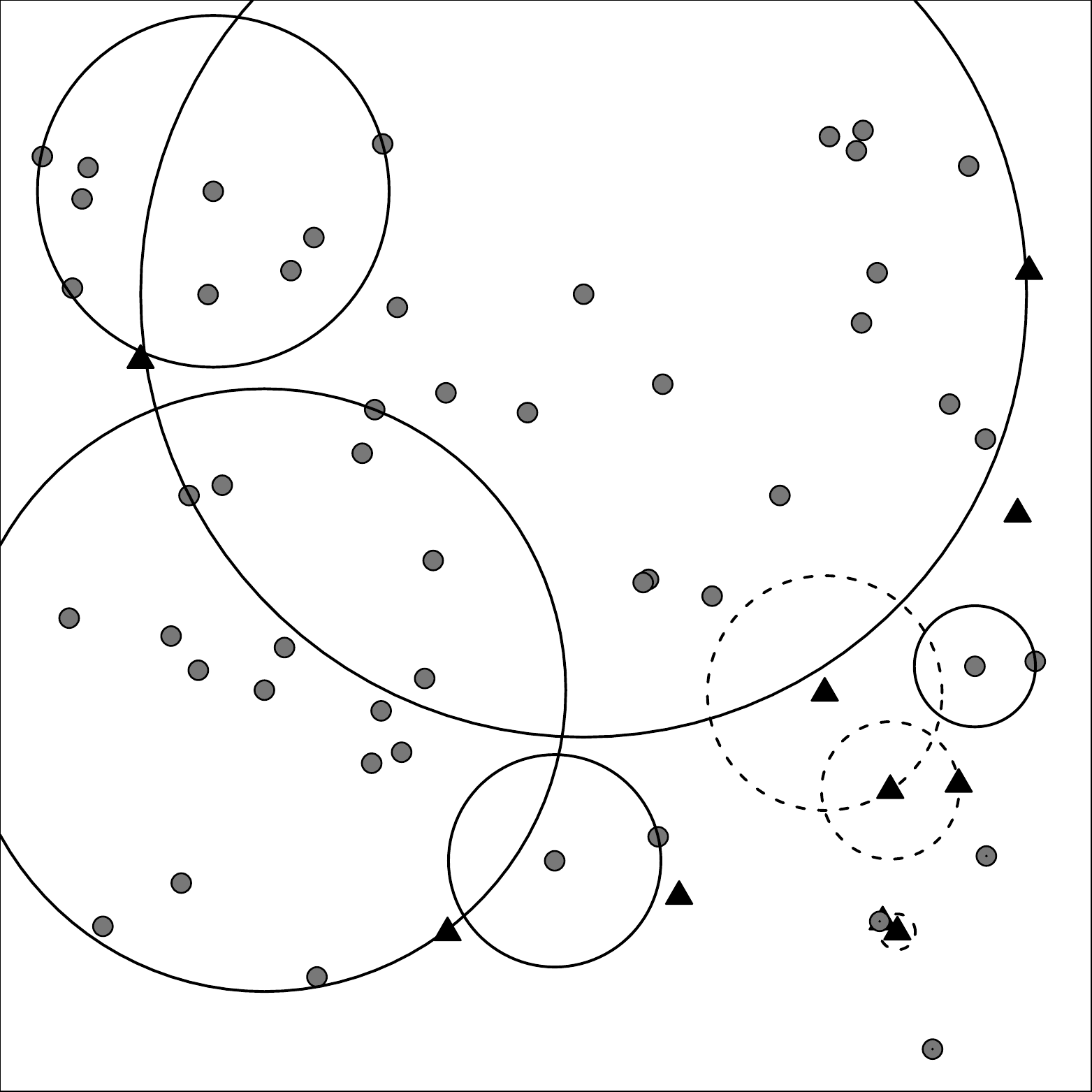} & \includegraphics[scale=0.40]{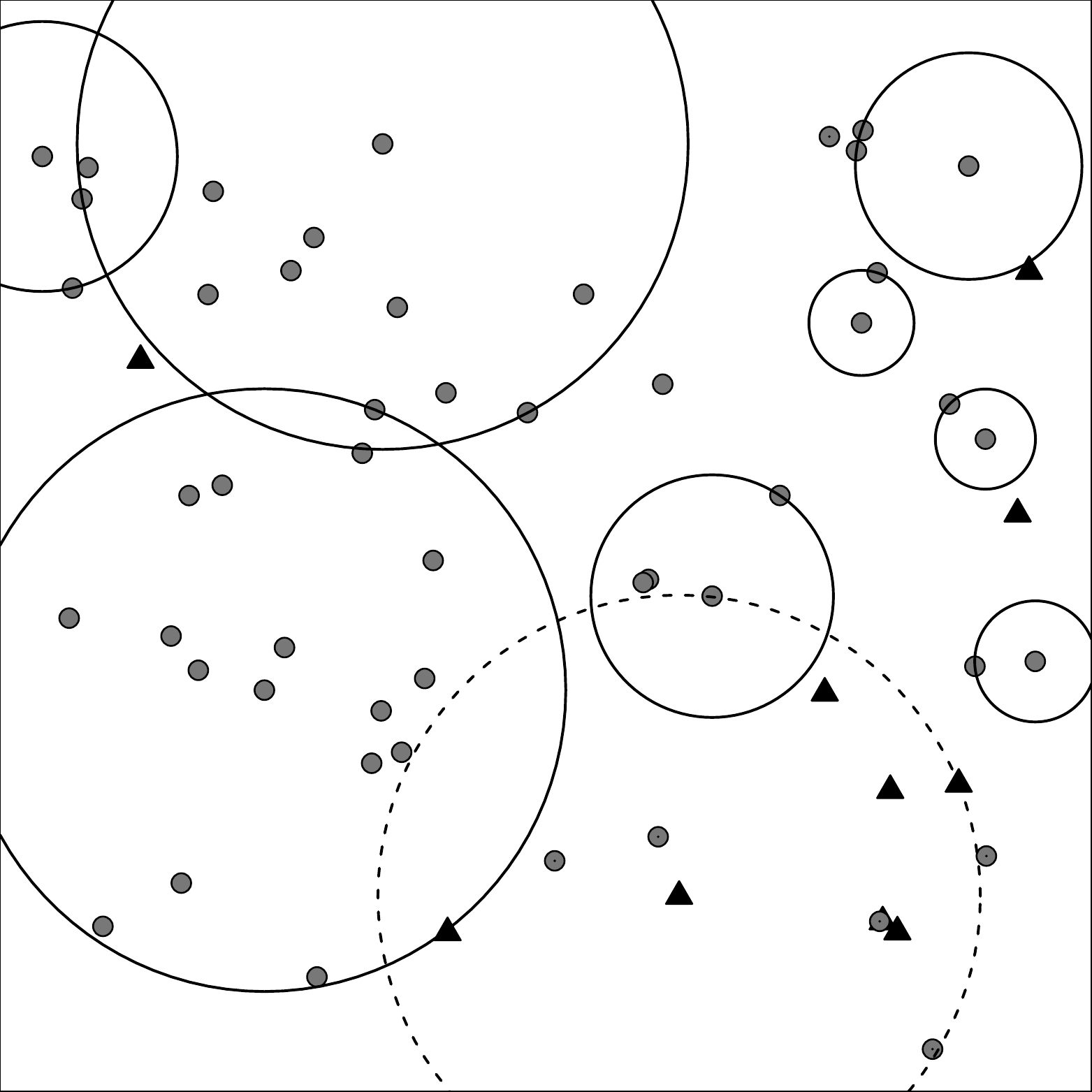} \\
\end{tabular}
\caption{An illustration of the covering balls associated with majority and minority P-CCCD (left) and the corresponding RW-CCCDs ($\tau=0.0001$) (right) of an imbalanced data set in a two-class setting where majority and minority class points are represented by grey dots and black triangles, respectively. }
\label{majvsmin}
\end{figure}

In Figure~\ref{majvsmin}, we illustrate the behavior of balls associated with P-CCCDs and RW-CCCDs.
Note that in both families of digraphs,
balls of the majority class tend to be larger and hence are more likely to catch more majority class points. Since the majority class has much more members than the minority class, balls of the majority points are more likely to catch the neighboring majority points. CCCD classifiers keep the information of ball centers and their associated radii. Larger cardinality of the majority class allows the construction of bigger balls and hence, larger values of radii are more likely to correspond to larger number of catched class members. As a result, CCCDs balance the data set and, at the same time, preserve the information of the local density by retaining the radii. The data set becomes balanced since the center of balls are the points of the new training data set which will be employed later in classification.
	
The loss of information in undersampling schemes are of course inevitable, however it is possible to preserve a portion of that discarded information by other means. EasyEnsemble is an ensemble classifier used for that very purpose; however, it needs multiple classifiers to be employed. Each classifier is trained on a different balanced subset of the original training data set, and hence the ensemble classifier preserves the information on the entire data set given by a collection of unbiased classifiers. On the other hand, CCCDs achieve the same goal by transforming the density around points into the radii. 
CCCDs resemble cluster-based resampling methods in that regard. Instead of randomly sampling the data set, 
cluster-based resampling schemes divide each class into clusters, and then, oversample the minority class or undersample the majority class proportional to each subclass size.
Covering balls of CCCDs have a similar purpose which has also been discussed in \citet{priebe2003dna}. They use the covering balls of the minimum dominating sets to explore the latent subclasses of each class of gene expression data sets. In fact, the balls of CCCDs may correspond to clusters. Hence, sets of points associated with each cluster is undersampled to a single point (i.e., a prototype or a dominating point), and the information on the cluster is provided by the radius which represents the density of that cluster. The bigger the radius, the more influence a prototype has over the domain. In P-CCCDs, the radii may be sensitive to noise, but RW-CCCDs ignore noisy points to avoid overfitting. Moreover, in RW-CCCDs, we have an additional statistic provided by each cluster, the score given in Equation~(\ref{rwscore}) based on the random walk. We use both the radii and these scores to define the RW-CCCD classifiers, and thus achieve better performing classifiers with more reduction and less information loss.
	
We approach the problem of class imbalance from the perspective of class overlapping problem as well. Several researchers on class imbalance revealed that overlap between the class supports degrade the classification performance of imbalanced data sets even more \citep[see][]{prati2004,batista2004,batista2005,galar2012}. Let $E \subset \mathbb{R}^d$
be the overlapping region of these two class supports, $E:=s(F_X) \cap s(F_Y)$. Moreover, let $q(E):= |\mathcal{Y}_m \cap E|/|\mathcal{X}_n \cap E|$ be ratio of class sizes restricted to the region $E \subset \mathbb{R}^d$. We say $q(E)$ is the ``local" imbalance ratio with respect to $E$. Also, let the ``global" imbalance ratio be $q=q(\mathbb{R}^d)=m/n$. Throughout this work, in both simulated and real data examples, we study and discuss the local imbalance ratio $q(E)$ restricted to the overlapping region $E$ and the global imbalance ratio $q$. We specifically illustrate the performance of several classifiers for various levels of class imbalance (local or global) and class overlapping, and assess the performance of CCCD classifiers compared to weak and strong versions of $k$-NN, SVM and C4.5 classifiers.

\section{Comparing CCCDs with Other Classifiers} \label{comparecccd}

We study the performance of CCCD classifiers in comparison with weak and strong classifiers in two separate sections.
Recall that we call a classifier ``weak" when the method is inherently sensitive to class imbalance,
and ``strong" when it is non-sensitive (or less sensitive). We use the area under curve (AUC) measure to evaluate the performance of the classifiers on the imbalanced data sets \citep{lopez2013}. AUC measure is often used on imbalanced real data classes. This measure has been shown to be better than the correct classification rate in general \citep{huang2005}. We discuss the computational complexity of weak classifiers to emphasize the testing speed of CCCD classifiers when trained by imbalanced data sets. Finally, we compare both weak and strong classifiers with CCCDs on real data sets by considering the overlapping and imbalance ratios of all data sets.

\subsection{Monte Carlo Simulation Study with Weak Classifiers} \label{simdata}

In this section, we compare the CCCD-based classifiers, namely P-CCCD and RW-CCCD, with $k$-NN, support vector machines (SVM) and C4.5, on simulated data sets. These classifiers are listed in Table~\ref{methods}.
We employ the \texttt{cccd}, \texttt{e0171} and \texttt{RWeka} packages in R
to classify test data sets with the P-CCCD, SVM (with Gaussian kernel) and C4.5 classifiers, respectively \citep{marchette2013,e10712014,Rteam}.

For each of four classification methods other than C4.5,
we assign the optimum parameter values which are the best performing values among all considered parameters. For example, an optimum P-CCCD parameter $\tau$ is found in a preliminary (pilot) Monte Carlo simulation study associated with the main simulation setting (i.e., the same setting of the main simulation).
In the pilot study, we perform a Monte Carlo simulation with 200 replications and
count how many times a $\tau$ value has the maximum AUC among $\tau=0.0,0.1,\cdots,1.0$ in 200 trials.
Note that, since $\tau \in (0,1]$, we denote $\tau=\epsilon$ (machine epsilon) as $\tau=0$ for the sake of simplicity. For each replication of the pilot simulation, we (i) classify the test data set with all $\tau$ values, (ii) record the $\tau$ values with maximum AUC and (iii) update the count of the recorded $\tau$ values.
Finally, we appoint the one that has the maximum count (i.e., the mode) as the $\tau^*$, the best performing or the optimum $\tau$.
Then, we use $\tau^*$ as the parameter of P-CCCD classifier in our main simulation.
The parameters of optimal $k$-NN, SVM and RW-CCCD classifiers are defined similarly.
SVM methods often incorporate both a kernel parameter $\gamma$ and a constrained violation cost $C$.
We only optimize $\gamma$ since the selection of an optimum $C$ parameter will be more crucial for cost-sensitive SVM methods.
Moreover, we consider two versions of the C4.5 classifier where both incorporate Laplace smoothing. The first tree classifier, C45-LP, prunes the decision tree with \%25 confidence level but the second classifier, C45-LNP, does not prune at all.

\begin{figure}
\centering
\includegraphics[scale=0.8]{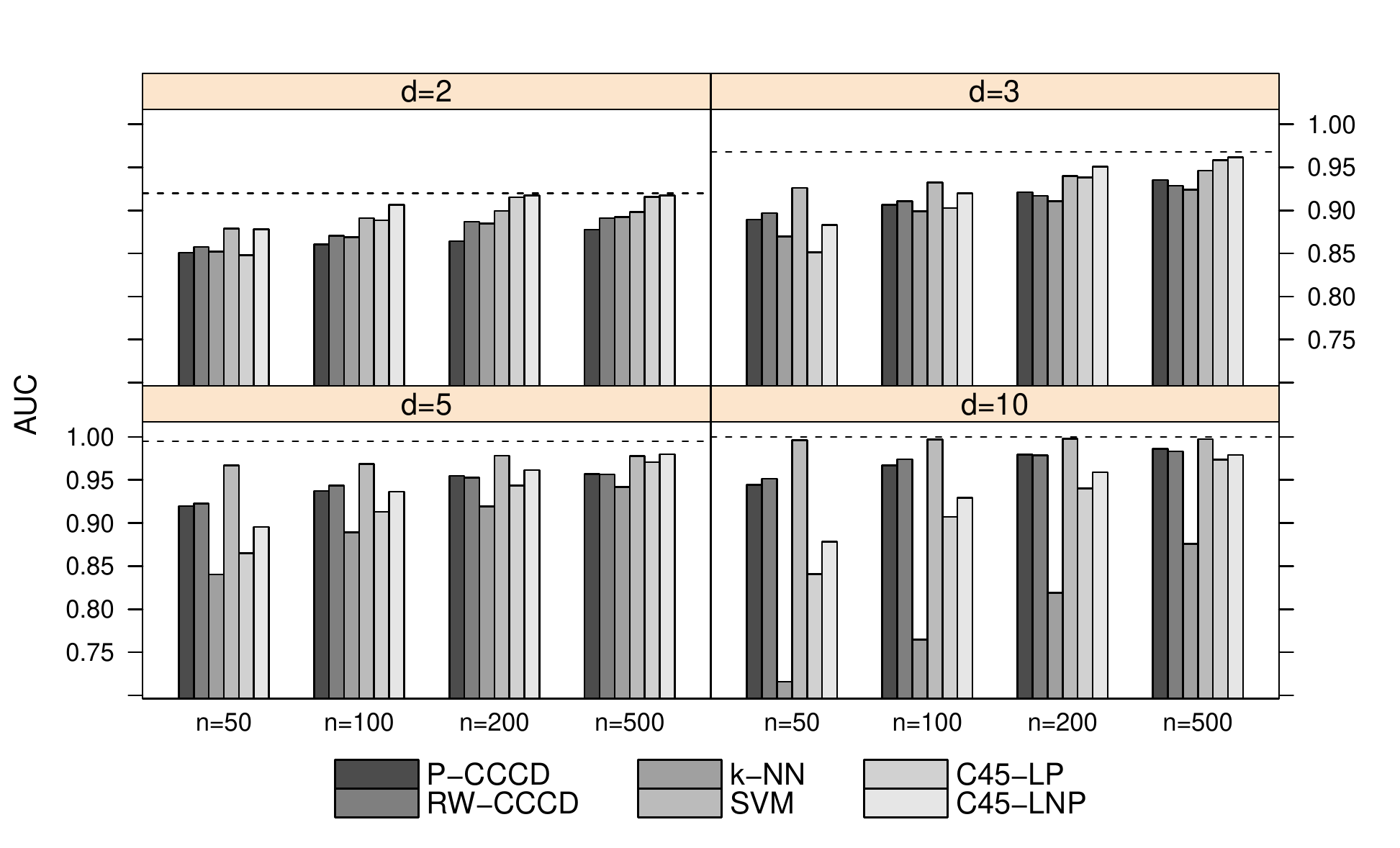}
\caption{AUCs in the two-class setting, $F_X =  U(0,1)^d$ and $F_Y =  U(0.3,0.7)^d$
under various simulation settings, with $d=2,3,5,10$ and equal class sizes $m=n=50,100,200,500$.}
\label{embed}
\end{figure}

\begin{figure}
\centering
\includegraphics[scale=0.6]{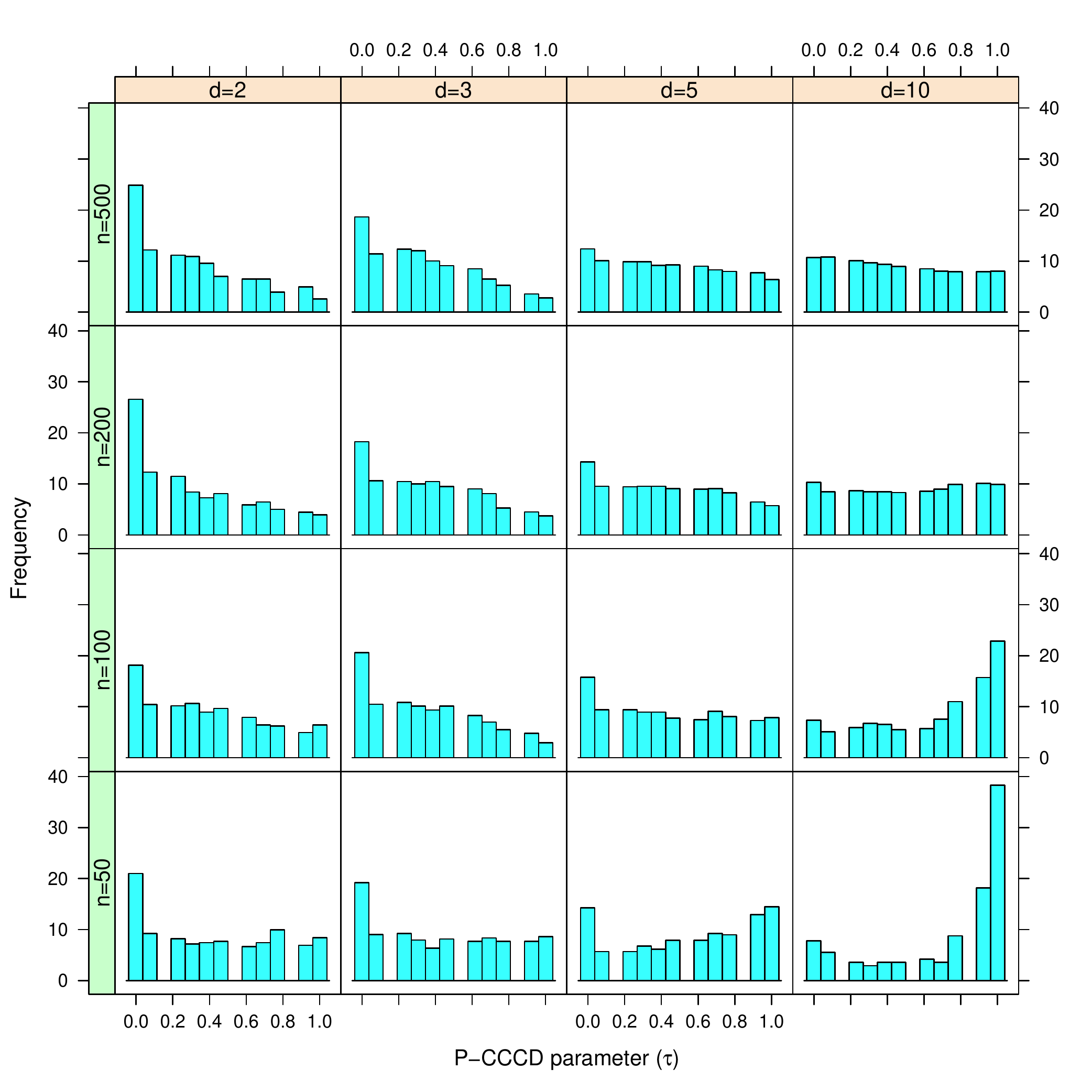}
\caption{Frequencies of the best performing $\tau$ values among $\tau=0.0,0.1,\cdots,1.0$ in our pilot study (this is used to determine the optimal $\tau$ used in P-CCCD).
The simulation setting is same as to the one presented in Figure~\ref{embed}.}
\label{paramIR1}
\end{figure}

\begin{figure}
\centering
\includegraphics[scale=0.6]{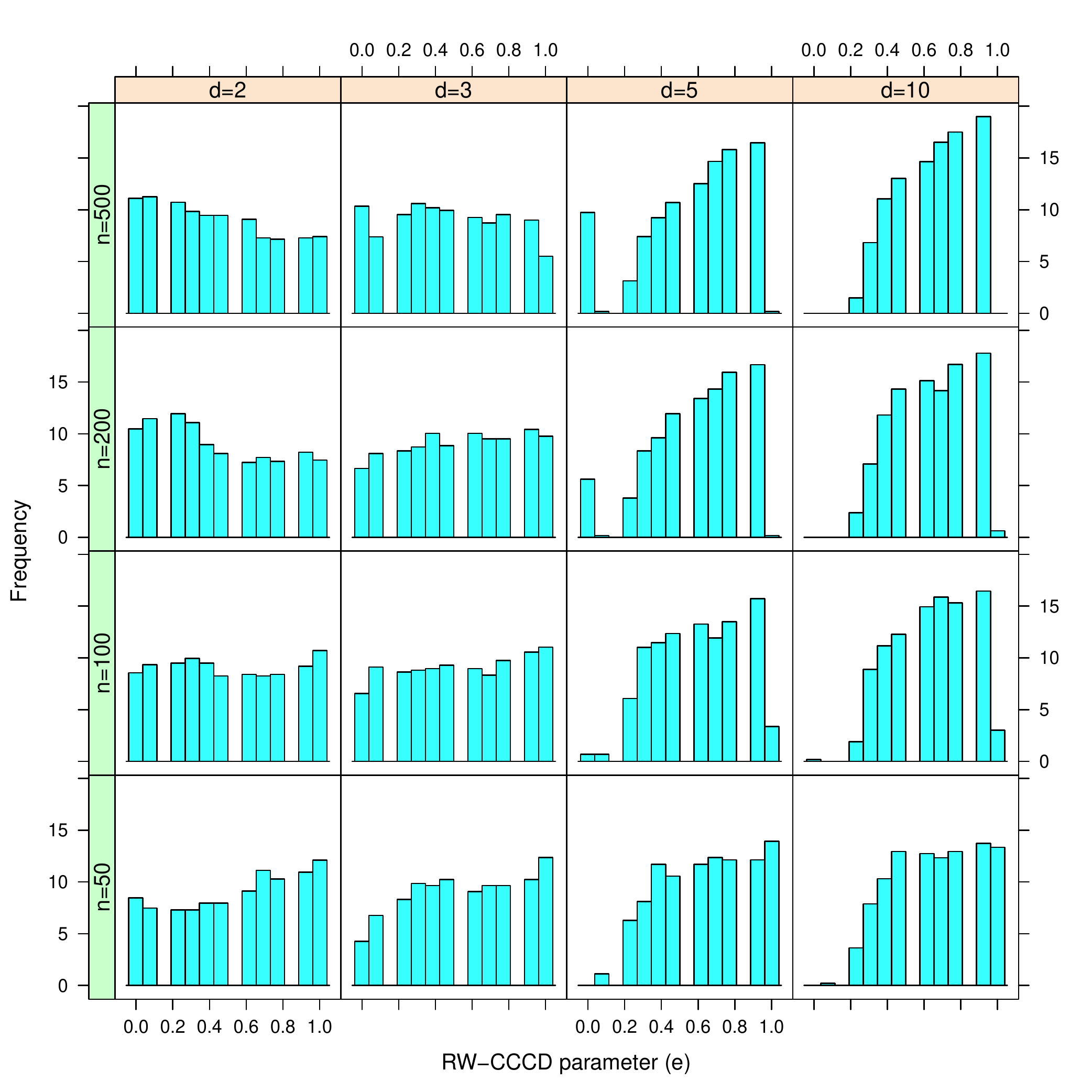}
\caption{Frequencies of the best performing $e$ values among $e=0.0,0.1,\cdots,1.0$ in our pilot study (this is used to determine the optimal $e$ used in RW-CCCD).
The simulation setting is same as to the one presented in Figure~\ref{embed}.}
\label{paramIR2}
\end{figure}

We first consider a simulation setting similar to the one in \citet{devinney2002} where CCCD classifiers showed relatively good performance compared to the $k$-NN classifier. Here, we simulate a two-class setting where observations from both classes are drawn from separate multivariate uniform distributions: $F_X =  U(0,1)^d$ and $F_Y =  U(0.3,0.7)^d$ for $d = 2,3,5,10$. Notice that $s(F_Y) \subset s(F_X)$; i.e., $E=s(F_Y)$. We perform Monte Carlo replications where on each replication, we train the data with equal sizes of observations ($m=n$) from each class for $n = 50,100,200,500$. On each replication, we record the AUC measures of the classifiers on the test data set with 100 observations from each class, resulting a test data set of size 200. We simulate test data sets until AUCs of all classifiers achieve a standard error below 0.0005. Average of AUCs of all classifiers in Table~\ref{methods} are given in Figure~\ref{embed} for all $(n,d)$ combinations. Additionally, in Figure~\ref{paramIR1}, we report the $\tau$ values of best performing P-CCCD classifiers in our pilot simulation study for all $(n,d)$ combinations. In Figure~\ref{paramIR1}, there are separate histograms for each combination. Each histogram represents
the number of times a $\tau$ value has the maximum AUC. Also in Figure~\ref{paramIR2}, we report the $e$ values of the best performing RW-CCCD classifiers of the same pilot simulation study for $e = 0,0.1,\cdots,1.0$.
	
\begin{table}
\centering
\scriptsize
\begin{tabular}{l|p{9cm}}
Method & Description \\
\hline
P-CCCD &  P-CCCD with the optimum $\tau$ (in the pilot study) among $\tau = 0,0.1,\cdots,1.0$ \\
RW-CCCD &  RW-CCCD with the optimum $e$ (in the pilot study) among $e = 0,0.1,\cdots,1.0$ \\
$k$-NN & $k$-NN with optimum $k$ (in the pilot study) among $k = 1, 2, \cdots, 30$ \\
SVM & SVM with the radial basis function (Gaussian) kernel with the optimum $\gamma$ (in the pilot study) among $\gamma= 0.1, 0.2, \cdots, 3.9, 4.0$ \citep{joachims1999} \\
C45-LP & C4.5 with Laplace smoothing and reduced error pruning (\%25 confidence) \\
C45-LNP & C4.5 with Laplace smoothing and no pruning \\
\hline
\end{tabular}
\caption{The description of classifiers employed in the article.}
\label{methods}
\end{table}

We start by investigating the effect of $\tau$ and $e$ on CCCD classifiers. The relationship between $\tau$, $n$ and $d$ can also be observed in Figure~\ref{paramIR1}. The higher the $\tau$ value, the better the performance of P-CCCD classifier with increasing $d$ and decreasing $n$. This may indicate that balls with $\tau=0$ (i.e., $\tau=\epsilon$) represent the density around their centers better for low dimensional data sets. However, with increasing dimensionality and lower class sizes, the set of points gets sparser in $\mathbb{R}^d$. In the case of RW-CCCD, classifiers with high $e$ values are either better or comparable to those with lower $e$ values. The scores $T_x$ of covering balls are definitely beneficial to the performance of the RW-CCCD classifiers, however with increasing $n$ and decreasing $d$ (especially for $n=500$ and $d=2$) RW-CCCD with lower $e$ is better since the radii successfully represent the density around the prototype points due to the high number of observations in the data set.
	
Figure~\ref{embed} illustrates the AUCs of all classifiers along with the Bayes optimal performance given with the dashed line. Comparing the performance of CCCD classifiers with other classification methods, we observe that RW-CCCD and P-CCCD classifiers outperform the $k$-NN classifier when the support of one class is entirely embedded inside that of the other class. These results are similar to the conclusions of \citet{devinney2002}: with increasing dimensionality, the difference between $k$-NN and CCCD classifiers becomes more apparent, i.e., CCCD classifiers have nearly 0.20 AUC more than $k$-NN. On the other hand, the SVM classifier has about 0.05 more AUC than P-CCCD and RW-CCCD classifiers, especially for lower class sizes. Although, both versions of CCCD classifiers outperform the $k$-NN and C4.5 classifiers with increasing dimensionality, the gap between these two classifiers and CCCD classifier is getting narrower with increasing class sizes. The RW-CCCD classifier is slightly better than the P-CCCD classifier for lower $n$.
In addition, C45-LNP achieves slightly better results than C45-LP.
	
\begin{figure}
\centering
\includegraphics[scale=0.7]{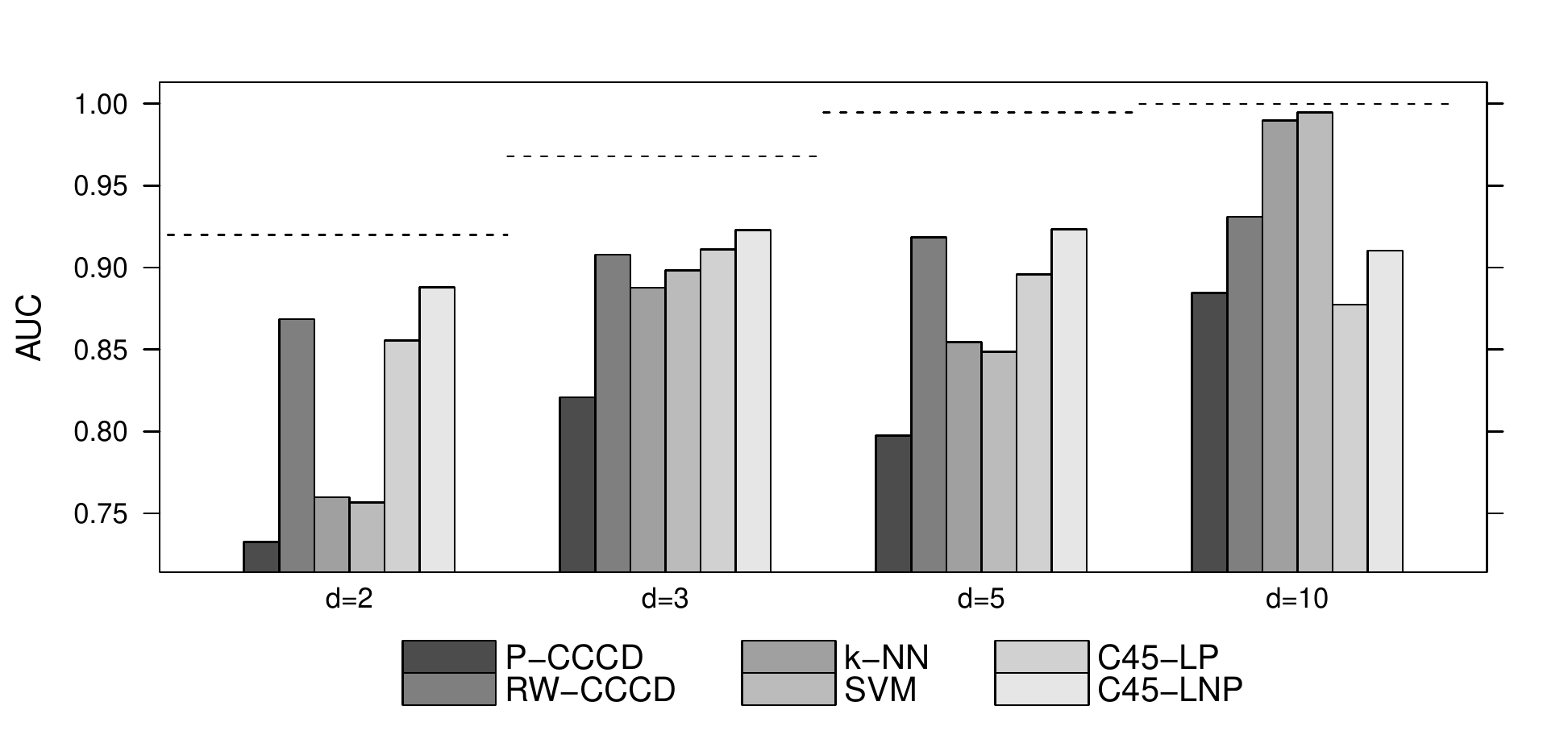}
\caption{AUCs in a two-class setting, $F_X =  U(0,1)^d$ and $F_Y =  U(0.3,0.7)^d$ with fixed $n=200$ and $m=50$ in $d=2,3$,
and with fixed $n=1000$ and $m=50$ in $d=5,10$.}
\label{embed2}
\end{figure}

In the setting presented in Figure~\ref{embed},
apparently, two classes overlap on the region $E = s(F_Y)=(0.3,0.7)^d$ which is the entire support of the class $\mathcal{Y}$.
For equal class sizes, $q=m/n=1$ but $q(E) \approx (1/0.4)^d = \Vol(s(F_X))/\Vol(s(F_Y))$,
where $\Vol(\cdot)$ is the volume functional.
The classes are clearly imbalanced in $E$, although $m=n$.
Hence, class $\mathcal{X}$ becomes the minority and class $\mathcal{Y}$ becomes the majority class with respect to $E$. However, readjusting the class sizes $m$ and $n$ might change the performance of P-CCCD and RW-CCCD classifiers compared to the $k$-NN and C4.5 classifiers. Therefore, we conduct another simulation study with classes from the same uniform distributions, but we set $m=50$ and $n=200$ for $d=2,3$, and $m=50$ and $n=1000$ for $d=5,10$. In this experiment, we simulated 4 times more $\mathcal{X}$ class members than $\mathcal{Y}$ for $d=2,3$, and 20 times more for $d=5,10$. Results of this second experiment is given in Figure~\ref{embed2}. $k$-NN and C4.5 classifiers outperform P-CCCD classifier in all $d$ cases and has comparable AUC with SVM. However, only for $d=2,5$, RW-CCCD classifier achieves considerably more or comparable AUC compared to other classifiers.
In this example, $k$-NN classifiers have nearly 0.05 more AUC than P-CCCDs,
and also RW-CCCDs have, in general, 0.05 more AUC than $k$-NN classifiers.
	
Results from Figures~\ref{embed} and~\ref{embed2} seem conflicting to each other, even though $E = s(F_Y)$.
In the simulation setting of Figure~\ref{embed2}, we draw more samples from the class $\mathcal{X}$ to balance the class sizes with respect to $E$.
In fact, the effect on the difference of AUCs between CCCD, $k$-NN and C4.5 classifiers depends heavily on the local class imbalance restricted to the overlapping region $E$. The classes in region $E$ are less imbalanced in setting of Figure~\ref{embed2} than in the setting of Figure~\ref{embed}. Observe that $q(E) \approx (1/0.4)^d/4$ when $(m,n)=(50,200)$, $q(E) \approx (1/0.4)^d/20$ when $(m,n)=(50,1000)$, and $q(E) \approx (1/0.4)^d$ in $(m,n)=(50,50)$. Hence, $d$ does also affect the balance between classes. With increasing $d$, the region $E$ gets smaller in volume compared to $s(F_X)$ and, as a result, fewer points of the class $\mathcal{X}$ falls in $E$. Thus, we need to draw more samples from $\mathcal{X}$ as dimensionality increases, in order to balance the classes with respect to $E$. These results suggest that, the more imbalanced the data set in overlapping region $E$, the worse the performance of $k$-NN and C4.5 classifiers while CCCD classifiers preserve their classification performance.
So, CCCD classifiers exhibit robustness (to the class imbalance problem).
On the other hand, in Figure~\ref{embed}, we observe that the AUC of $k$-NN classifier approaches to the AUC of CCCD classifiers with increasing class sizes. Because, when $q$ and $q(E)$ are fixed, the classification performance still depends on individual values of $n$ or $m$.
This result is in line with the results of \citet{japkowicz2002}
who reported that the effect of class imbalance on the classification performance diminishes
if both class sizes are sufficiently large.
Furthermore, SVM classifier performs better than all classifiers in Figure~\ref{embed},
and performs worse than RW-CCCD classifiers only for $d=2,5$ in Figure~\ref{embed2}.
This might be an indication that SVM classifier is also not affected by the local class imbalance with respect to $E$,
and performs usually better than both P-CCCD and RW-CCCD classifiers if the support of one class is inside the other.
For the C4.5 classifier, on the other hand,
it is known for quite some time that the pruning is detrimental for classifying imbalanced data sets \citep{cieslak2008}.
In any case, C45-LNP has more AUC than C45-LP in all simulation settings.
	
In a two-class setting with an overlapping region $E$, we should expect CCCD classifiers to outperform $k$-NN classifiers in cases of (global or local) class imbalance. Let $F_X =  U(0,1)^d$ and $F_Y =  U(\delta,1+\delta)^d$ for $\delta,q=0.05,0.10,\cdots,0.95,1.00$; $d=2,3,5,10$; $n=400$ and $m=qn$. Here, the shifting parameter $\delta$ controls the level of overlap.
The class supports get more overlapped with decreasing $\delta$. Since $E=(\delta,1)^d$ and the supports of both classes are unit boxes, observe that $q(E) \approx q$. The closer the value of $q$ to $1$, more balanced the classes are.
We aim to address the relationship between the classifiers for various combinations of overlapping and global class imbalance ratios.
	
\begin{figure}
\centering
\includegraphics[scale=0.48]{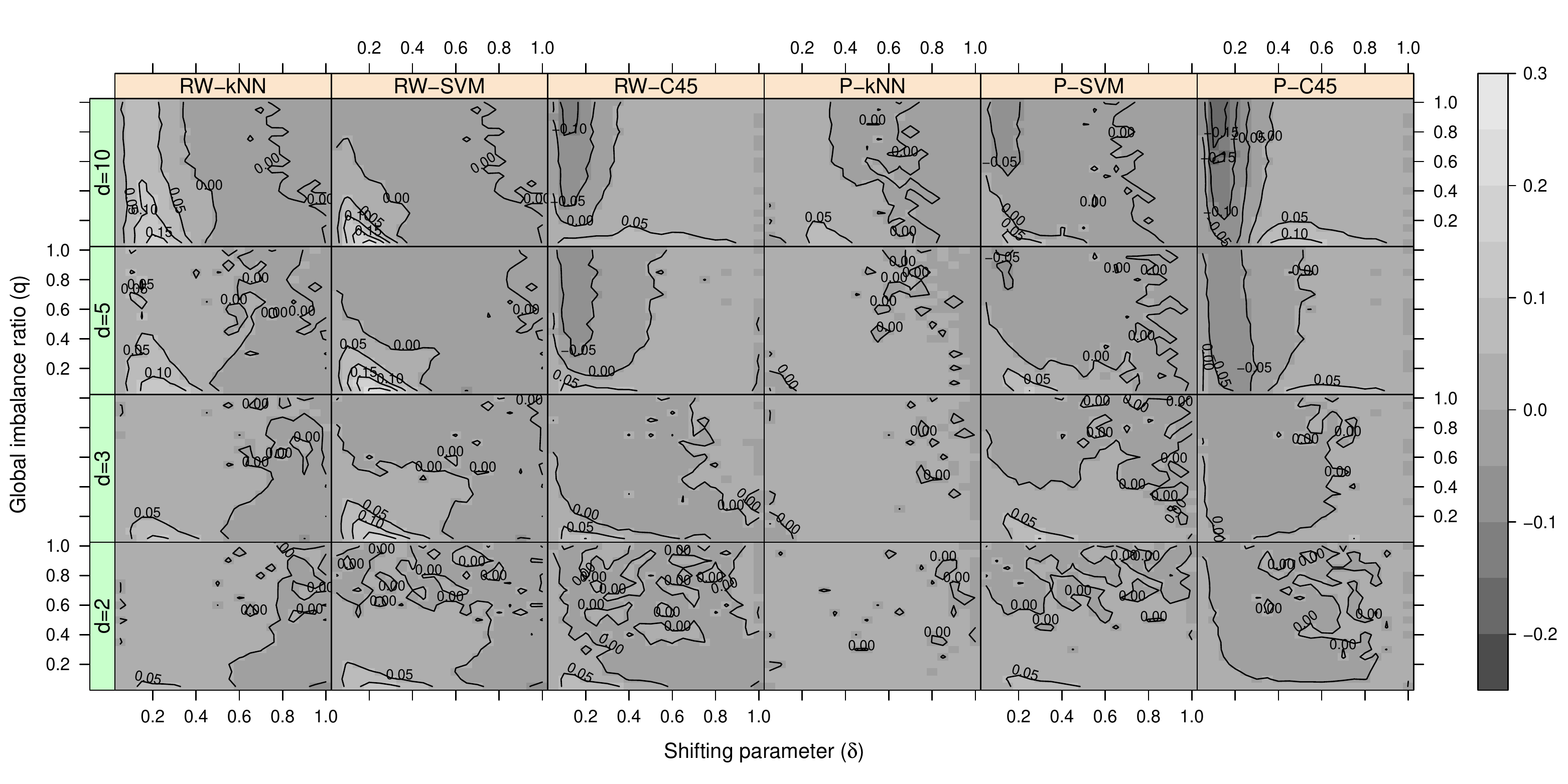}
\caption{Differences between the AUCs of CCCD and other classifiers.
For example, the panel titled with "RW-kNN" presents AUC(RW-CCCD)-AUC($k$NN).
In this two-class setting, classes are drawn from $F_X =  U(0,1)^d$ and $F_Y =  U(\delta,1+\delta)^d$ with $d=2,3,5,10$.
Each cell of the grey scale heat map corresponds to a single combination of simulation parameters
$\delta,q=0.05,0.1,\cdots,0.95,1.00$ with $n=400$ and $m=qn$.}
\label{shiftcccd}
\end{figure}

Figure~\ref{shiftcccd} illustrates the difference between AUCs of CCCD and other classifiers ($k$-NN, SVM and C4.5)
in separate heat maps for $d=2,3,5,10$.
We use the unpruned C4.5 classifier C45-LNP, since it tends to perform better for imbalanced data sets, and we refer to C45-LNP as C4.5 for simplicity.
Each cell of a single heat map is associated with a combination of $\delta$ and $q$ values.
Lighter tone cells indicate that CCCD classifiers are better than the other classifiers in terms of AUC,
and vice versa for the darker tones.
When the classes are imbalanced and moderately overlapping, RW-CCCD classifier has at least 0.05 more AUC than all other non-CCCD classifiers but P-CCCD classifier is only better than all others provided that $d=10$.
If the classes are balanced or their supports are not considerably overlapping, there seem to be no visible difference between CCCD and the other classifiers. Thus, the other classifiers suffer from the imbalance of the data while CCCD classifiers show robustness to the class imbalance. 
But more importantly, this difference is getting more emphasized with increasing dimensionality. 
When $d$ is high, fewer points of the minority class fall in $E$ although $q(E)$ is fixed.
When the classes are imbalanced, if the minority class have substantially small size, the class imbalance problem becomes more detrimental \citep{japkowicz2002}. Under the conditions that the data set has substantial imbalance and overlapping, AUC of RW-CCCD classifier is followed, in order by, the AUC of C4.5, SVM and $k$-NN classifiers.
	
Unlike the comparison of CCCD and SVM classifiers in Figures~\ref{embed} and \ref{embed2}, SVM classifier has less AUC than CCCD classifiers with low $\delta$ and low $q$ values in Figure~\ref{shiftcccd}. In this setting, $n$ is fixed to 400 and the lowest value of $m$ is 20. Compared to our experiments in Figures~\ref{embed} and \ref{embed2}, this setting produces highly imbalanced data sets (one class has far more observations than the other, $m << n$). \citet{akbani2004} conducted a detailed investigation and listed some reasons of SVM classifier being sensitive to highly imbalanced UCI data sets \citep{BacheLichman}. They did not, however, address the problem of overlapping class supports but offered a modification to SMOTE algorithm in order to improve the robustness of SVM. On the other hand, especially for $d=5$ and $d=10$, SVM, $k$-NN and C4.5 classifiers have more AUC than CCCD classifiers with increasing $q$ and decreasing $\delta$. This may indicate that other weak classifiers are better than CCCD classifiers for balanced classes.

\begin{figure}
\centering
\includegraphics[scale=0.55]{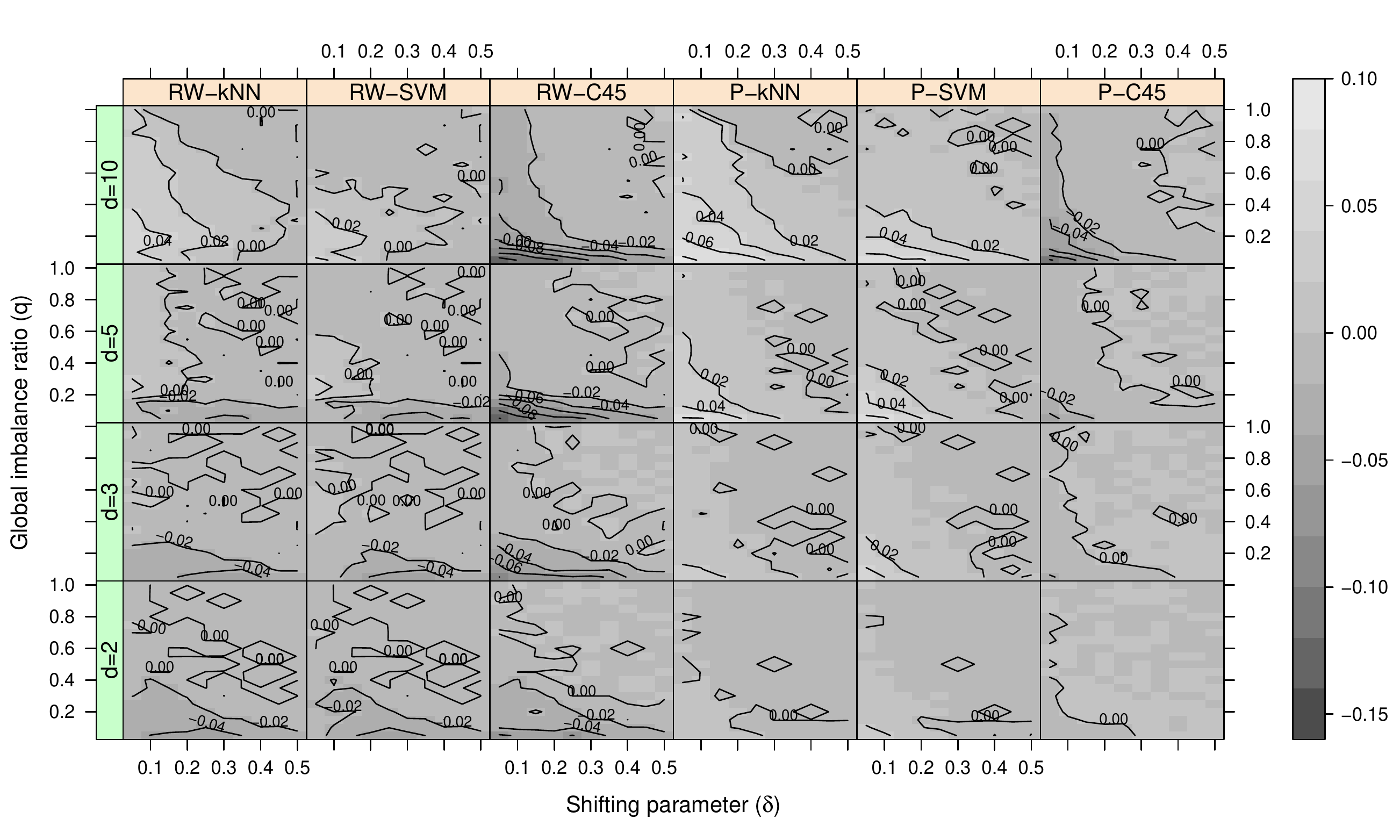}
\caption{Differences between the AUCs of CCCD and other classifiers (see Figure~\ref{shiftcccd} for details).
In this two-class setting, classes are drawn from $F_X =  U(0,1)^d$ and $F_Y =  U((1+\delta,2+\delta) \times (0,1)^{d-1})$ where $d=2,3,5,10$, $\delta=0.05,0.1,\cdots,0.45,0.50$ and $q=0.05,0.1,\cdots,0.95,1.00$ with $n=400$ and $m=qn$. AUCs of all classifiers are over $88\%$ since the class supports are well separated.}
\label{horizshiftcccd}
\end{figure}

The effects of class imbalance might also be observed when the class supports are well separated. If the class supports are disjoint,
that is $s(F_X) \cap s(F_Y)=\emptyset$, the AUC is fairly high. However, it might still be affected by the global imbalance level, $q$.
Therefore, we simulate a data set with two classes where
$F_X =  U(0,1)^d$ and $F_Y =  U((1+\delta,2+\delta) \times (0,1)^{d-1})$.
Figure~\ref{horizshiftcccd} illustrates the results of this simulation study. Both class supports are $d$ dimensional unit boxes as in the previous simulation setting, however they are now disjoint (separated along the first dimension). In addition, the parameter $\delta$ controls the smallest distance between the class supports where $\delta=0.05,0.10,\cdots,0.45,0.50$.
With increasing $\delta$, the points of class $\mathcal{Y}$ move further away from the points of $\mathcal{X}$.
Figure~\ref{horizshiftcccd} illustrates the difference between AUCs of CCCD and other classifiers under this simulation setting.
	
In Figure~\ref{horizshiftcccd}, unlike the performance of CCCD classifiers in Figure~\ref{shiftcccd}, P-CCCD classifiers have more AUC than RW-CCCD classifiers. When classes are imbalanced and supports are close, P-CCCD classifiers outperform both SVM and $k$-NN classifiers for all $d$ values, but RW-CCCD classifiers have nearly 0.03 more AUC than these classifiers only in $d=10$. However, this is not the case with C4.5 classifier since none of the classifiers outperform C4.5; that is, C4.5 yields over 0.04 more AUC than CCCD classifiers.
A well separated data set is more likely to be classified better with C4.5 tree classifier because a single separating line exists between the two class supports.
Hence, C4.5 locates such a line and efficiently classifies points regardless of the distance between class supports as long as the distance is positive.
On the other hand, the balls of P-CCCD classifiers tend to establish appealing covers for the class supports because the supports do not overlap.
P-CCCD classifiers establish covering balls, big enough to catch substantial amount of points from the same class.
Similarly, RW-CCCD classifiers establish pure covers, and this is the result of the separation between class supports.
However, P-CCCD classifiers achieve better classification performance than RW-CCCD classifiers.
When the classes are well separated,
the radii of a ball centered at $x$, say from class $\mathcal{X}$, in the random walk
is likely $\max_{z \in \mathcal{X}_n} d(z,x)$ but in P-CCCD classifiers,
it is $\min_{z \in \mathcal{Y}_n} d(z,x)$.
In fact, the RW-CCCD classifiers are nearly equivalent to P-CCCD classifiers.
Thus, when $\tau > 0$, P-CCCD classifiers are more likely to produce bigger balls than RW-CCCD classifiers, and potentially avoid overfitting.
	
In Figure~\ref{horizshiftcccd}, RW-CCCD classifiers have slightly or considerably less AUC than other classifiers when data sets are imbalanced and the supports are slightly far away from each other.
The random walk contaminates the class cover with some non-target class points to improve the classification performance.
However, since the classes are well separated and one class has substantially fewer points than the other,
random walks are likely to yield balls to cover some points from the support of the non-target class, resulting in a degradation in the performance of RW-CCCD classifiers. On the other hand, P-CCCD classifiers outperform both $k$-NN and SVM classifiers for lower $q$ and lower $\delta$. The closer and more imbalanced the data, the better the performance of P-CCCDs than other classifiers. Although the classes do not overlap, the effect of class imbalance is still observed when the supports are close. When there is mild imbalance between classes, CCCD classifiers have either comparable or less AUC. In addition, note that the performances of SVM and $k$-NN classifiers deteriorate but P-CCCD classifiers preserve their AUC with increasing $d$.
Let $E' \subset \mathbb{R}^d$ be some region that contains points of both classes which are sufficiently close to the decision boundary.
With increasing $d$, fewer minority class points are in this region,
and hence fewer members of this class fall in $E'$.
As a result, the performance of both SVM and $k$-NN classifiers suffer from local class imbalance with respect to $E'$.
	
Finally, we investigate the effect of dimensionality when classes are balanced (i.e., $q=1$) and their supports are overlapping. In this setting, $F_X =  U(0,1)^d$ and $F_Y =  U(\delta,1+\delta)^d$. Here, let $q(E) \approx q = 1$, hence the classes are also locally balanced with respect to $E$ as well as being globally balanced. Also, $\delta$ controls the level of overlap between two classes.
However, we define $\delta$ in such a way that the overlapping ratio $\alpha \in [0,1]$ is same for all dimensions
where overlapping ratio is defined as
\begin{equation} \label{alpha}
	\alpha = \frac{\Vol(s(F_\mathcal{X}) \cap s(F_\mathcal{Y}))}{\Vol(s(F_\mathcal{X}) \cup s(F_\mathcal{Y}))}=\frac{(1-\delta)^d}{2-(1-\delta)^d}.
\end{equation}
When $\alpha$ is 0, the supports are well separated, and when $\alpha$ is 1,
the supports of classes are the same, i.e.,
$s(F_X) = s(F_Y)$.
The closer $\alpha$ to 1, the more the supports overlap. Observe that $\delta \in [0,1]$ can be expressed in terms of the overlapping ratio $\alpha$ and dimensionality $d$ as
\begin{equation} \label{delta}
\delta = 1-\left(\frac{2\alpha}{1+\alpha}\right)^{1/d}.
\end{equation}

\begin{figure}
\centering
\includegraphics[scale=0.48]{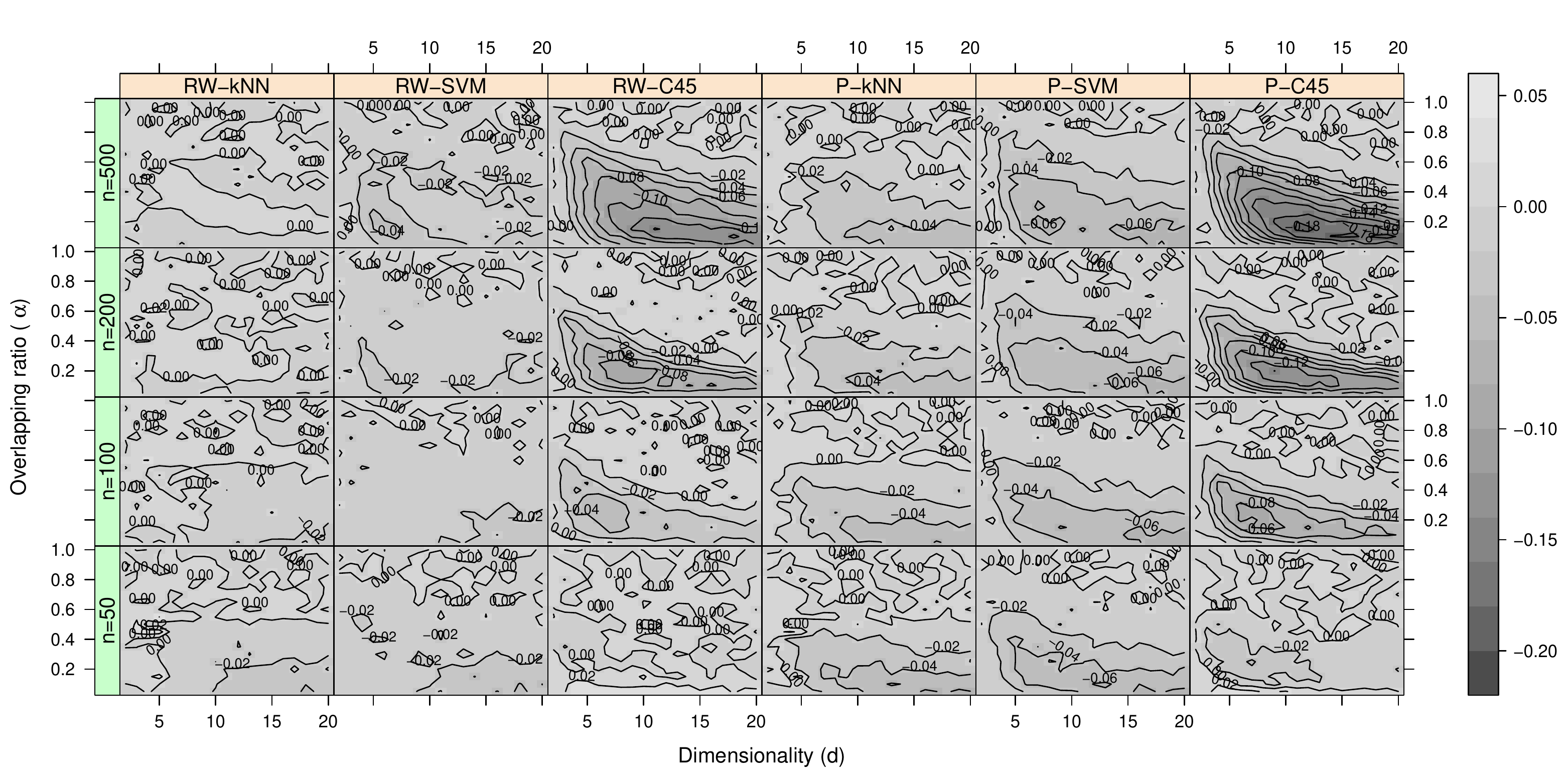}
\caption{Differences between the AUCs of CCCD and other classifiers (see Figure~\ref{shiftcccd} for details).
In this two-class setting, classes are drawn from $F_X =  U(0,1)^d$ and $F_Y =  U(\delta,1+\delta)^d$ where $n=50,100,200,500$, $\alpha=0.05,0.1,\cdots,0.45,1.00$ and $d=2,3,4,\cdots,20$. }
\label{dimcccd}
\end{figure}

\noindent
Hence, we calculate $\delta$ for each $(d,\alpha)$ combination by the Equation~(\ref{delta}).
In Figure~\ref{dimcccd}, each cell of the grey scale heat map corresponds to a single combination of simulation parameters $\alpha=0.05,0.1,\cdots,0.95,1.00$ and $d=2,3,4,\cdots,20$.
In Figure~\ref{dimcccd}, the differences between the AUCs of CCCD classifiers and other classifiers are up to 0.20.
The $k$-NN and SVM classifiers have comparable performance with CCCD classifiers, or outperform both CCCD classifiers.
However, C4.5 has more AUC with increasing $d$.
Employing CCCD classifiers do not considerably increase the classification performance over other classifiers when classes are balanced.

\subsection{Empirical Comparison of CCCD-based and Strong Classifiers}

In this section, we compare the CCCD-based classifiers with strong versions of $k$-NN, SVM and C4.5 classifiers on simulated data sets. Each classifier is modified in three different schemes, namely, resampling, ensemble and cost-sensitive schemes.
We use SMOTE+ENN algorithm as the resampling scheme and EasyEnsemble algorithm as the ensembling scheme.
As for the cost-sensitive versions of the weak classifiers, we adjust the classifiers into recognizing class weights.
For $k$-NN, we employ an algorithm giving more weight on neighboring minority class members;
for SVM, we use two separate constrained violation costs for each corresponding class; and for C4.5, we employ the C5.0.
With three schemes and three weak classifiers, we get nine strong classifiers to study.
We list and describe all these schemes in Table~\ref{methodsred}.

\begin{table}
\centering
\scriptsize
\begin{tabular}{l|p{9cm}}
Method & Description \\
\hline
SMOTE+ENN & A combination of SMOTE ($t=2$ and $k=5$) and ENN ($k=3$) \citep{batista2004}.\\
EasyEnsemble & A combination of undersampling ($T=4$) and Adaboost ($s_i=10$) for $i=1,2,\cdots,T$ \citep{liu2009}\\
C5.0 & The cost-sensitive version of C4.5 \citep{kuhn2013}.\\
C$k$NN & A cost-sensitive version of $k$-NN \citep{barandela2003}.\\
CSVM & A cost-sensitive version of SVM \citep{chang2011}.\\
\hline
\end{tabular}
\caption{The description of classifiers employed in the article.}
\label{methodsred}
\end{table}

SMOTE+ENN algorithm, first, oversamples the entire training data set by generating artificial points in between a point and its neighbors. Specifically, for each point in the data set, $t$ points among $k$ neighbors are selected, and until the data set is balanced, new artificial points are generated in between these points and their selected neighbors. Later, ENN algorithm cleans the data set of noisy points by checking all points if the majority of their $k$ neighbors are labeled as the class of the point. If not, the point is erased from the data set. Simply, SMOTE+ENN is a hybrid of over and undersampling methods.
EasyEnsemble algorithm is a hybrid of undersampling and ensemble methods. An ensemble of weak classifiers  is established by generating $T$ many undersampled balanced data sets from the training data set.
Then, each data set is used to train individual Adaboost classifiers with $s_i$ many weak classifiers
for $i=1,2,\ldots,T$.
Hence, EasyEnsemble is an ensemble of $\sum_{i=1}^T s_i$ many weak classifiers.

\begin{figure}
\centering
\includegraphics[scale=0.50]{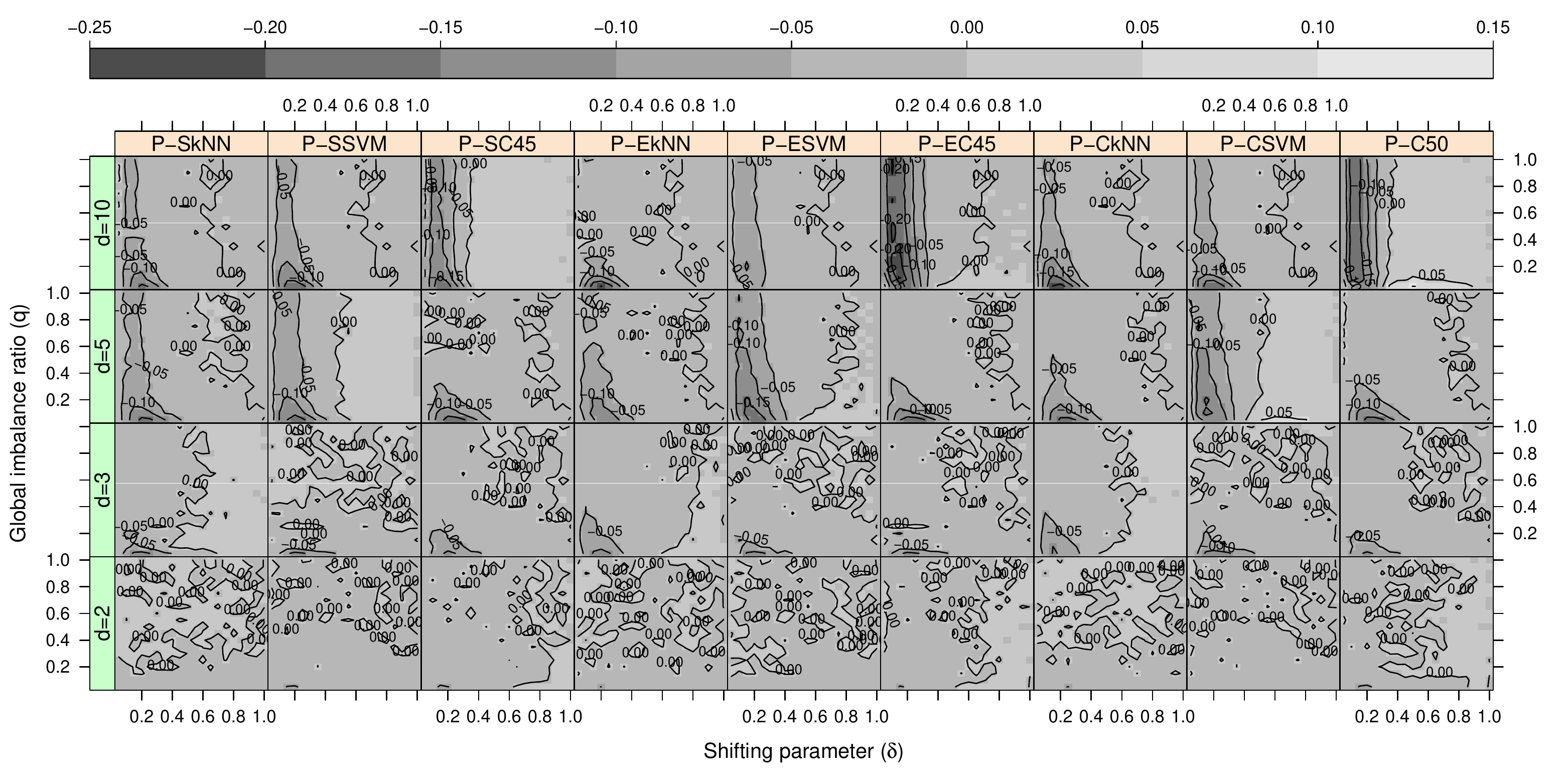}
\includegraphics[scale=0.50]{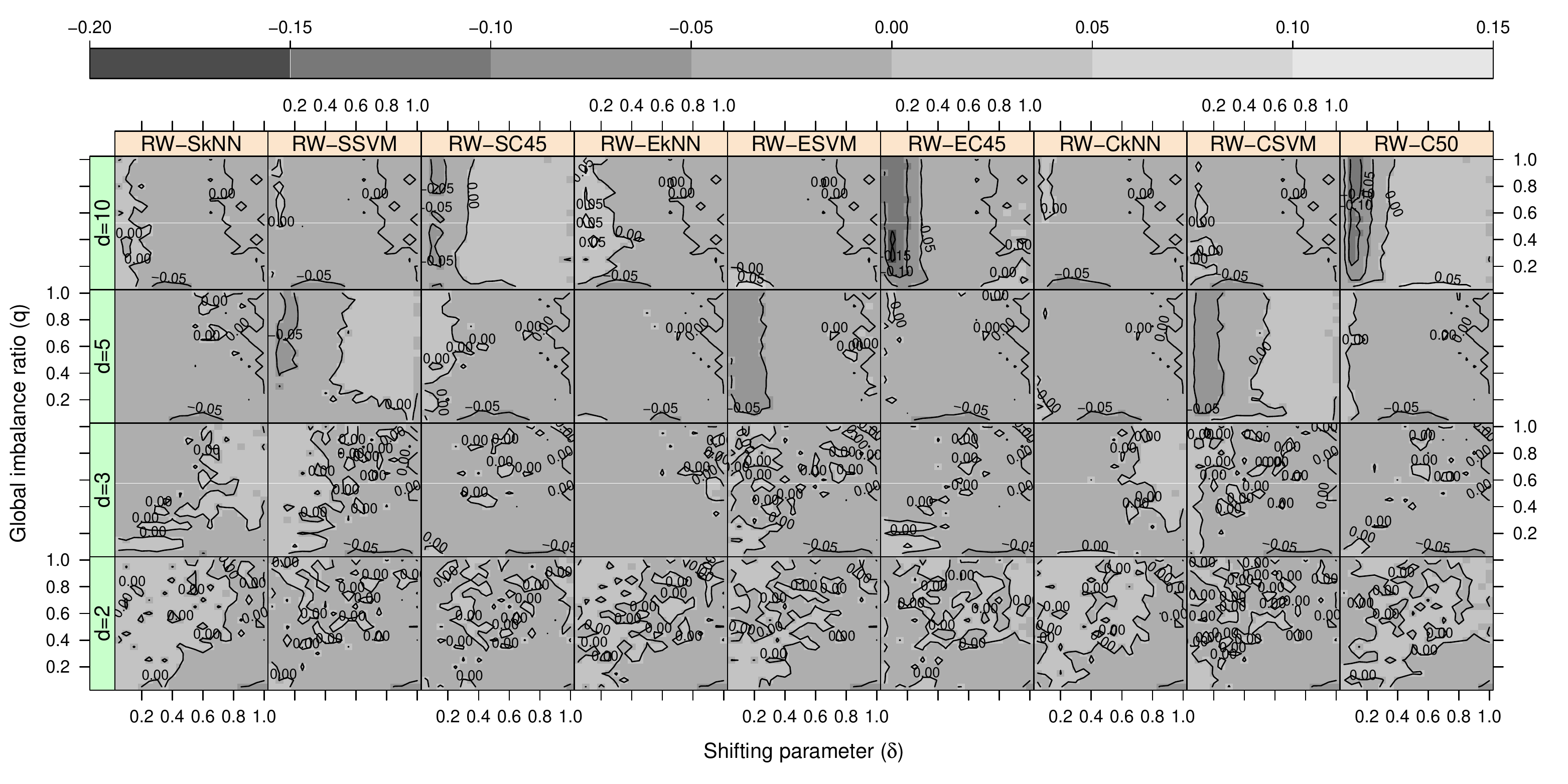}
\caption{Differences between the AUCs of CCCD and other classifiers (see Figure~\ref{shiftcccd} for details). P-CCCD in (top) and RW-CCCD in (bottom). Here, resampling scheme strong classifiers are coded with ``S", ensemble schemes with ``E", and cost-sensitive schemes with ``C". For example, ``SkNN" refers to the resampling schemed $k$-NN classifier. In this two-class setting, classes are drawn from $F_X =  U(0,1)^d$ and $F_Y =  U(\delta,1+\delta)^d$ where $d=2,3,5,10$, $q,\delta=0.05,0.1,\cdots,0.95,1.00$ with $n=400$ and $m=qn$.}
\label{shiftred}
\end{figure}

We choose one of the simulation settings conducted in Section~\ref{simdata}. Since CCCD classifiers are observed to be better than other classifiers when both class imbalance and overlapping occurred, we only compare CCCD classifiers with strong classifiers on a single simulation setting.
Hence we choose the setting presented in Figure \ref{shiftcccd}, i.e., we let $F_X =  U(0,1)^d$ and $F_Y =  U(\delta,1+\delta)^d$ for $\delta,q=0.05,0.10,\cdots,0.95,1.00$, $n=400$ and $m=qn$. We aim to highlight the differences between the strong classifiers and CCCD classifiers for various combinations of overlapping and class imbalance ratios.
The results on average AUCs of each strong classifier is given in Figure~\ref{shiftred}.
In general, RW-CCCDs seem to perform better than P-CCCDs.
For $d>2$, P-CCCDs have nearly 0.10 less AUC than RW-CCCDs when the classes are substantially overlapping and imbalanced, and it is observed that P-CCCDs are usually worse compared to the strong classifiers considered.
However, the AUCs of RW-CCCD classifiers are either comparable or slightly less compared to others
with the most difference being seen in the case of $d=10$ when RW-CCCDs compared to EC4.5 and  C5.0,
ensemble and cost-sensitive versions of the C4.5 classifier, respectively.
However, with decreasing $\delta$ and $q$, the RW-CCCDs have only 0.05 less AUC than others.
Also, RW-CCCDs seem to have 0.05 more AUC than C5.0 for moderately overlapping and imbalanced data sets, and seem to have 0.05 more AUC then ESVM, ensemble based SVMs, when the data set is both overlapping and imbalanced.
This suggests that RW-CCCDs yield comparable results in comparison to the state-of-the-art robust methods when class imbalance and class overlapping co-exist.
Additionally, we show in Section~\ref{sec:complex} that RW-CCCDs generate prototype sets that considerably reduce the training data set.

\subsection{Complexity Analysis of the Classifiers} \label{sec:complex}

In Table~\ref{complex}, we compare training and testing time and space complexities of P-CCCDs, RW-CCCDs, $k$-NN, SVM and C4.5 classifiers. Let $N=n+m$ be the size of training data set. C4.5 is the fastest among all classifiers and requires the least space. However, unpruned C4.5 constitute a tree with its space complexity increasing exponentially on $d$ since the data set is divided into at most two for all dimensions. The remaining classifiers are all instance based learning methods which depend on a matrix of distances between the points of training data set. Hence, their space complexity is at least $\mathcal{O}(N^2)$ and they run in at least $\mathcal{O}(N^2d)$ time. Both SVM and RW-CCCD classifiers run in $\mathcal{O}(N^3)$ time for $d < N$, and P-CCCD runs in $\mathcal{O}(N^2d)$ time. Minimum dominating set problem of P-CCCDs are polynomial time reducible to minimum set cover problems, and hence they run in $\mathcal{O}(N^2)$ time in the worst case but they require the computation of the distance matrix which takes the most time. However, in RW-CCCDs, covering balls are re-defined each time a new point is added to the prototype set. As a result, this operation requires an additional sweep on the training set on each iteration which makes RW-CCCD run in $\mathcal{O}(N^3)$ time, for $d<N$.
For SVM, the training time of usual optimization algorithms is $\mathcal{O}(N^3)$ for $d<N$.
However, it is possible to reduce the complexity to $\mathcal{O}(N^{2.3})$
with sequential minimal optimization (SMO) method \citep{chang2011}.

\begin{table}
\centering
\begin{tabular}{ccccc}
& \multicolumn{2}{c}{Training} & \multicolumn{2}{c}{Testing} \\
\hline
& Time & Space & Time & Space \\
\hline
P-CCCD  & $\mathcal{O}(N^2d)$ & $\mathcal{O}(N^2)$ & $\mathcal{O}(Md)$ & $\mathcal{O}(Md)$ \\
RW-CCCD & $\mathcal{O}(N^3 + N^2d)$ & $\mathcal{O}(N^2)$ & $\mathcal{O}(Md)$ & $\mathcal{O}(Md)$ \\
$k$-NN  & $\cdots$ & $\cdots$ &  $\mathcal{O}(Md)$ & $\mathcal{O}(Md)$ \\
SVM     & $\mathcal{O}(N^3 + N^2d)$ & $\mathcal{O}(N^2)$ & $\mathcal{O}(Md)$ &$\mathcal{O}(Md)$ \\
C45-LNP & $\mathcal{O}(Nd^2)$ & $\mathcal{O}(Nd)$ & $\mathcal{O}(d)$ & $\mathcal{O}(2^d)$ \\
\hline
\end{tabular}
\caption{Training and testing space and time complexities of the weak classifiers.
$N$: size of training data, $M$: size of test data.}
\label{complex}
\end{table}

\begin{figure}
\centering
\includegraphics[scale=0.55]{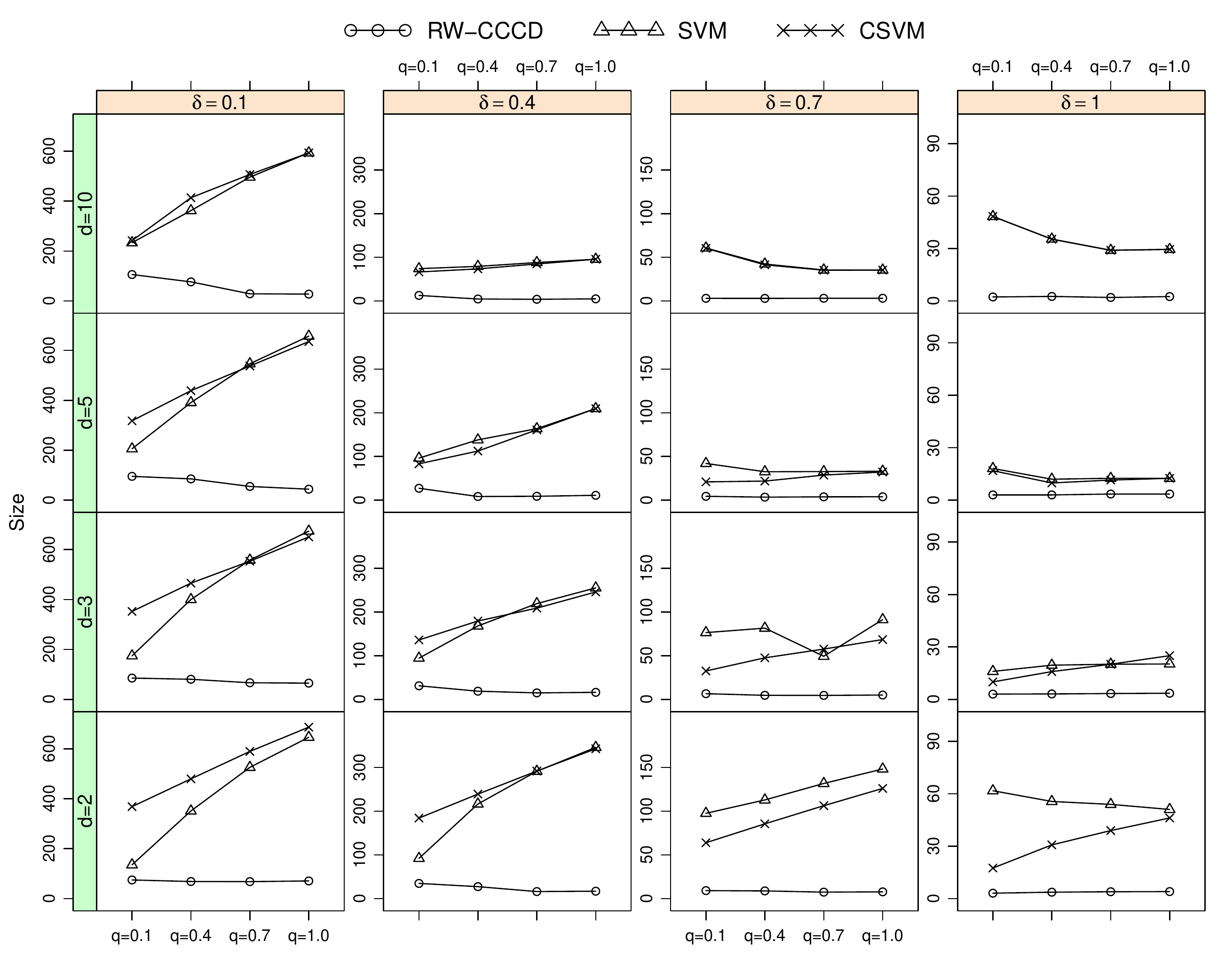}
\caption{Comparison of the sizes of reduced data sets in RW-CCCDs, SVM and CSVM classifiers.
Here ``size" refers to the number of covering balls in RW-CCCD or the number of support vectors in SVM classifiers.
In this two-class setting, classes are drawn from $F_X =  U(0,1)^d$ and $F_Y =  U(\delta,1+\delta)^d$ where $\delta,q=0.1,0.4,0.7,1.0$ with $n=400$ and $m=qn$.}
\label{reducesvm}
\end{figure}

Note that, $k$-NN does not require any training time or space,
and should use the entire training data set to classify the test data set.
However, CCCD and SVM classifiers reduce the training data set by means of prototype sets (minimum dominating sets in CCCD and support vectors in SVMs) even though their worst case testing space complexity is $\mathcal{O}(Nd)$. The entire training data set could be chosen as the prototype set for some cases, but we show that the data set is substantially reduced when the classes are imbalanced. In Figure~\ref{reducesvm}, we compare the sizes of the set of prototypes in RW-CCCDs and the set of support vectors in SVM and CSVM classifiers. We consider the simulation settings with two classes for $F_X =  U(0,1)^d$ and $F_Y =  U(\delta,1+\delta)^d$, $\delta,q=0.1,0.4,0.7,1.0$, $d=2,3,5,10$, $n=400$ and $m=qn$.

\begin{figure}
\centering
\includegraphics[scale=0.55]{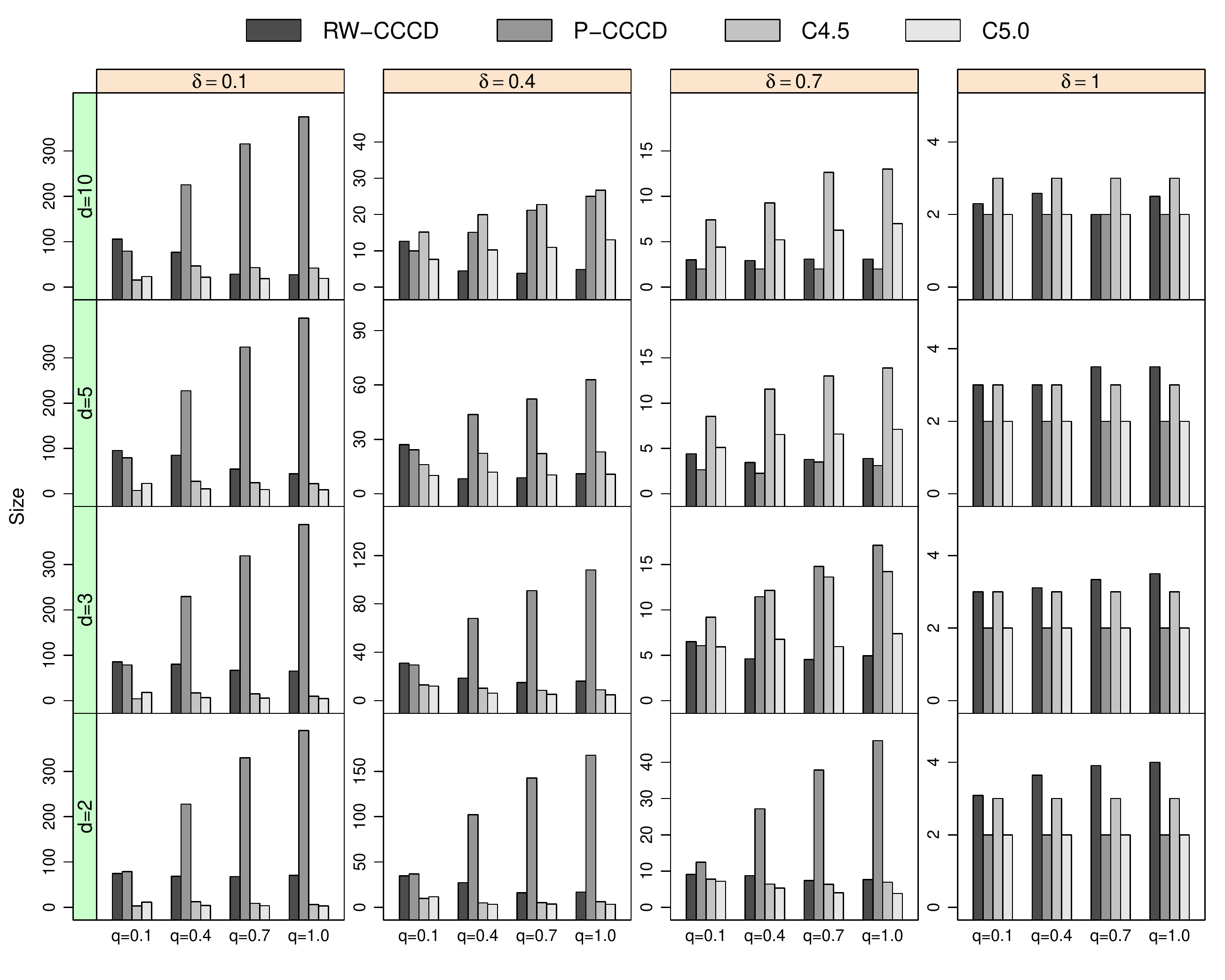}
\caption{Comparison of the sizes of reduced data sets in CCCDs and C4.5.
Here ``size" refers to the number of covering balls in CCCDs or the number of nodes in the decision tree of C4.5 classifiers.
The simulation setting is same as in Figure~\ref{reducesvm}.}
\label{reduceall}
\end{figure}

In Figure~\ref{reducesvm}, the number of both support vectors and prototypes decrease with increasing $\delta$. The prototype set heavily depends on the overlapping ratio between class supports.
Obviously, when points of either class are further away from each other, covering balls get bigger for CCCDs, and the separating hyperplane requires less support vectors. On the other hand, observe that the number of support vectors are much higher than the number of prototypes of RW-CCCDs. The number of support vectors decreases with decreasing $q$.
The more imbalanced the data set, the fewer support vectors are generated.
But in any case, RW-CCCDs still reduce the training data set more than SVMs.
In Figure~\ref{reduceall}, we compare the number of prototypes in both CCCD classifier and the size of the C4.5 and C5.0 classifier trees for the same simulation setting.

The number of prototypes in P-CCCDs are, in general, much higher than that of other classifiers.
Also, notice that the less imbalanced the classes are,
the less the data reduction in P-CCCDs. However, there is not much change in the number in RW-CCCDs, C4.5 and C5.0, and since the size of trees grows exponentially on $d$, the size of trees get bigger than the size of CCCDs for some substantially high $\delta$ and $d$. Moreover, the size of trees in C5.0 is considerably less than that in C4.5 \citep{kuhn2013}. Although the number of prototypes are much higher than the size of trees in highly overlapped and imbalanced cases, RW-CCCDs reduce the training set substantially more than C4.5 and C5.0 in moderately imbalanced and moderately overlapped higher dimensional settings.

\subsection{Real Data Examples}

In this section, we compare the performance of CCCD classifiers and all other weak and strong classifiers on several data sets from UC Irvine (UCI) Machine Learning and KEEL repositories \citep{BacheLichman,fdez2011}. To test the difference between the AUC of classifiers, we employ the 5x2 cross validation (CV) paired $t$-test \citep[see][]{dietterich1998} and the combined 5x2 CV $F$-test \citep[see][]{alpaydm1999}. The 5x2 CV test has been devised by \citet{dietterich1998} and found to be the most powerful test among those with acceptable type-I error. However, the test statistics of 5x2 $t$-tests depend on which one of the ten folds is used. Hence, \citet{alpaydm1999} offered a combined 5x2 CV $F$-test which works as an omnibus test for all ten possible 5x2 $t$-tests (for each five repetitions there are two folds, hence ten folds in total). Basically, if a majority of ten 5x2 $t$-tests suggests that two classifiers are significantly different in terms of performance, the $F$-test also suggests a significant difference. Hence, an $F$-test with high $p$-value suggests that most of the ten $t$-tests fail to have low $p$-values.

We also provide the overlapping ratios and imbalance levels of these data sets.
In a simulation study such as the ones in Section~\ref{simdata},
we have control on the overlapping region of two classes since we can choose the supports of the classes,
hence their overlapping region is exactly known.
However, in real data sets where the support of classes are neither defined nor available, we need methods to estimate the supports and hence estimate the overlapping ratios for the two classes. We employ the support vector data description (SVDD) method of \citet{tax2004} for this purpose. The method finds a description (or a region) of a data set, which covers a desired percentage of the points. SVDDs are also used in novelty or outlier detection. It has been inspired by the SVM classifiers and is based on defining a sphere around the data set. Similar to SVM, kernel functions can be employed to define more relaxed regions. SVDD is also a one-class learning method where the goal is to decide if a new point belongs to this particular class or not \citep{juszczak2002}. By using SVDD approach, \citet{xiong2010} found the SVDD regions of each class and its overlapping region. We also use SVDD to find the overlapping region $E$ of each pair and report on the imbalance ratio with respect to $E$. The overlapping ratio is the percentage of points from both classes that reside in $E$. We use the Ddtools toolbox \citep{Ddtools2014} of MATLAB environment to produce the SVDDs of classes. Our choice of the kernel is the same as we have used with SVM classifiers in this study, the radial basis (i.e., Gaussian) kernel; for consistency. However, the selection of $\sigma$ in the kernel is crucial for the SVDD region.
	
In Table~\ref{UCI}, we present the overlapping ratios and the imbalance in the overlapping areas of all data sets for $ \sigma = 2,3,\cdots,10$. Although the value of $\sigma$ produces different overlapping ratios, it is apparent that classes of data sets Ionosphere, Abalone19, Yeast4, Yeast6 and Yeast1289vs7 have more overlap than others, and these overlapping data sets have substantial local class imbalance in their respective overlapping regions. Other data sets have almost no overlapping nor imbalance in the overlapping regions even though their classes are globally imbalanced. One of these data sets is Yeast5 which has a imbalance ratio of $q=32.70$ but has no imbalance in the small overlapping region.

\begin{table}[t]
\centering
\scriptsize
\resizebox{\textwidth}{!}{
\begin{tabular}{cccccccccccccc}
Data & $q=m/n$ & $N$ & $d$ & & $\sigma=2$ & $\sigma=3$ & $\sigma=4$ & $\sigma=5$ & $\sigma=6$ & $\sigma=7$ & $\sigma=8$ & $\sigma=9$ & $\sigma=10$ \\
\hline \rule{0pt}{3ex}
\multirow{2}{*}{Sonar} & \multirow{2}{*}{1.14} & \multirow{2}{*}{208} & \multirow{2}{*}{61} & OR & 4\% & 19\% & 23\% & 25\% & 26\% & 26\% & 27\% & 28\% & 28\% \tabularnewline
 & & & & IR & 1.22 & 1.04 & 0.96 & 0.93 & 0.96 & 0.97 & 0.97 & 0.90 & 0.90\tabularnewline
 \hline \rule{0pt}{3ex}
 \multirow{2}{*}{Ionosphere} & \multirow{2}{*}{1.78} & \multirow{2}{*}{351} & \multirow{2}{*}{35} & OR & 25\% & 36\% & 66\% & 69\% & 66\% & 79\% & 61\% & 76\% & 81\% \tabularnewline
 & & & & IR & 90.00 & 62.50 & 8.70 & 6.20 & 5.44 & 3.67 & 5.02 & 3.98 & 3.31\tabularnewline
\hline \rule{0pt}{3ex}
 \multirow{2}{*}{Segment0} & \multirow{2}{*}{6.02} & \multirow{2}{*}{2308} & \multirow{2}{*}{20}  & OR & 0\% & 0\% & 0\% & 0\% & 0\% & 0\% & 0\% & 0\% & 0\% \tabularnewline
 & & & & IR & NA & NA & NA & NA & NA & NA & NA & NA & NA\tabularnewline
\hline \rule{0pt}{3ex}
 \multirow{2}{*}{Page-Blocks0} & \multirow{2}{*}{8.79} & \multirow{2}{*}{5472} & \multirow{2}{*}{11} & OR & 0.6\% & 0.6\% & 0.6\% & 0.8\% & 0.9\% & 1\% & 1\% & 1\% & 1\% \tabularnewline
 & & & & IR & 0.47 & 0.22 & 0.22 & 0.25 & 0.40 & 0.39 & 0.43 & 0.54 & 0.63\tabularnewline
\hline \rule{0pt}{3ex}
\multirow{2}{*}{Vowel0} & \multirow{2}{*}{9.98} & \multirow{2}{*}{988} & \multirow{2}{*}{14} & OR & 0\% & 0\% & 0\% & 0\% & 0\% & 0\% & 0\% & 0\% & 0\% \tabularnewline
 & & & & IR & NA & NA & NA & NA & NA & NA & NA & NA & NA\tabularnewline
\hline \rule{0pt}{3ex}
 \multirow{2}{*}{Shuttle0vs4} & \multirow{2}{*}{13.87} & \multirow{2}{*}{1829} & \multirow{2}{*}{10}  & OR & 0\% & 0\% & 0\% & 0\% & 0\% & 0\% & 0\% & 0\% & 0\% \tabularnewline
 & & & & IR & NA & NA & NA & NA & NA & NA & NA & NA & NA\tabularnewline
\hline \rule{0pt}{3ex}
\multirow{2}{*}{Yeast4} & \multirow{2}{*}{28.10} & \multirow{2}{*}{1484} & \multirow{2}{*}{9}  & OR & 45\% & 27\% & 39\% & 37\% & 26\% & 26\% & 26\% & 26\% & 31\% \tabularnewline
 & & & & IR & 18.97 & 99.75 & 18.50 & 18.24 & 392.00 & 390.00 & 391.00 & 390.00 & 41.18\tabularnewline
 \hline \rule{0pt}{3ex}
\multirow{2}{*}{Yeast1289vs7} & \multirow{2}{*}{30.70} & \multirow{2}{*}{947} & \multirow{2}{*}{9} & OR & 45\% & 44\% & 69\% & 43\% & 30\% & 29\% & 29\% & 29\% & 29\% \tabularnewline
 & & & & IR & 24.47 & 23.76 & 24.34 & 23.00 & 47.33 & 45.83 & 45.83 & 45.66 & 45.50\tabularnewline
 \hline \rule{0pt}{3ex}
 \multirow{2}{*}{Yeast5} & \multirow{2}{*}{32.70} & \multirow{2}{*}{1484} & \multirow{2}{*}{9} & OR & 6\% & 3\% & 0\% & 0\% & 0\% & 0\% & 6\% & 7\% & 0.1\% \tabularnewline
 & & & & IR & NA & 1.15 & NA & NA & NA & NA & NA & NA & NA\tabularnewline
   \hline \rule{0pt}{3ex}
\multirow{2}{*}{Yeast6} & \multirow{2}{*}{41.40} & \multirow{2}{*}{1484} & \multirow{2}{*}{9}  & OR & 30\% & 46\% & 42\% & 31\% & 30\% & 30\% & 10\% & 13\% & 13\% \tabularnewline
 & & & & IR & 64.14 & 21.03 & 27.59 & 76.33 & 73.66 & 73.66 & 38.25 & 63.00 & 63.00\tabularnewline
   \hline \rule{0pt}{3ex}
 \multirow{2}{*}{Abalone19} & \multirow{2}{*}{129.40} & \multirow{2}{*}{4174} & \multirow{2}{*}{9} & OR & 25\% & 20\% & 15\% & 14\% & 13\% & 12\% & 12\% & 11\% & 11\% \tabularnewline
 & & & & IR & 104.30 & 104.30 & 163.75 & 197.33 & 278.00 & 262.50 & 253.00 & 244.00 & 234.00\tabularnewline
\hline
\end{tabular}
}
\caption{Overlapping ratios and (local) imbalance ratios in the overlapping region of data sets.
``IR" stands for the imbalance ratio in the overlapping region and
``OR" stands for the overlapping ratio which is the percentage of points from both classes residing in the overlapping region. IR=``NA" indicates that one of the classes has no members in the intersections of SVDD regions of classes.}
\label{UCI}
\end{table}

In Table~\ref{52test}, we give the average AUC measures and their standard deviations of all CCCD-based and other classifiers according to the 5x2 CV scheme for the data sets. All other classifiers, weak or strong, have been two-way tested with 5x2 CV $F$-test against both RW-CCCD and P-CCCD classifiers. Their $p$-values are also provided in Table~\ref{52test}. For each of five repetitions, we divide the data into two folds. The AUC of fold 1 is given by using fold 1 as a training set and fold 2 as the test set. For fold 2, the process is similar. We repeated these experiments five times for all three classifiers. Looking at results from 11 data sets, RW-CCCD usually performs better than P-CCCD classifiers, and in addition, ensemble based classifiers perform the best in general. Moreover, ensemble classifiers seem to perform better than RW-CCCDs but this difference is usually not significant, meaning RW-CCCDs perform comparable to ensemble classifiers in more than few folds of all ten folds. For example, compared to ensemble methods, RW-CCCD has nearly 0.07 less AUC in Yeast5, 0.02 less AUC in Yeast6, and 0.1 less AUC in Yeast1289vs7 data set. The difference is significant, however, with the data set Abalone19 with a level of $<0.03$. Although RW-CCCD achieves an average AUC value $0.6$, ensemble classifiers achieve over $0.7$. On the other hand, there is no significant difference between AUCs of RW-CCCD and ensembles in other highly overlapped and locally imbalanced data sets. On these data sets, RW-CCCD have significantly more AUC than weak classifiers and have AUC comparable to strong classifiers. Thus, these results from real data sets seem to resonate with the results from our simulations and further support the robustness of CCCD classifiers to the class imbalance problem.

\begin{table}
\centering
\resizebox{\textwidth}{0.85\textheight}{
\begin{sideways}
\begin{tabular}{ccccccccccccc}
&& Ionosphere & Sonar &	Yeast6 & Yeast5 & Yeast4 & Yeast1289vs7 & Vowel0 & Shuttle0vs4 & Abalone19 & Segment0 & Page-Blocks0 \\
\hline \rule{0pt}{3ex}
RW-CCCD & AUC & 0.917$\mp$0.023 & 0.722$\mp$0.050 & \textbf{0.866$\mp$0.051} & \textbf{0.898$\mp$0.063} & \textbf{0.807$\mp$0.048} & \textbf{0.643$\mp$0.057} & 0.877$\mp$0.046 & \textbf{0.996$\mp$0.003} & \textbf{0.603$\mp$0.065} & 0.895$\mp$0.011 & \textbf{0.875$\mp$0.019} \\
P-CCCD  & AUC & \textbf{0.934$\mp$0.032} & \textbf{0.805$\mp$0.045} & 0.755$\mp$0.053 & 0.793$\mp$0.094 & 0.602$\mp$0.051 & 0.556$\mp$0.038 & \textbf{0.958$\mp$0.025} & 0.988$\mp$0.016 & 0.506$\mp$0.019 & \textbf{0.957$\mp$0.010} & 0.869$\mp$0.009 \\
\hline \rule{0pt}{3ex}
\multirow{3}{*}{$k$-NN} & AUC & 0.803$\mp$0.019 & \textbf{0.804$\mp$0.027} & 0.786$\mp$0.032 & 0.839$\mp$0.063 & 0.619$\mp$0.038 & 0.562$\mp$0.04 & 0.971$\mp$0.031 & 0.996$\mp$0.004 & 0.512$\mp$0.015 & \textbf{0.988$\mp$0.004} & 0.863$\mp$0.009 \\
&$p$-value (vs RW)& 0.000 & 0.107 & 0.072 & 0.573 & 0.009 & 0.087 & 0.037 & 0.693 & 0.104 & 0.000 & 0.604 \\
&$p$-value (vs P)& 0.005 & \textbf{0.473} & 0.552 & 0.209 & 0.650 & 0.177 & 0.504 & 0.534 & 0.735 & \textbf{0.026} & 0.600 \\
\hline \rule{0pt}{3ex}
\multirow{3}{*}{SVM} & AUC & \textbf{0.949$\mp$0.010} & 0.719$\mp$0.057 & 0.710$\mp$0.045 & 0.741$\mp$0.069 & 0.527$\mp$0.024 & 0.507$\mp$0.014 & 0.956$\mp$0.039 & 0.984$\mp$0.012 & 0.500$\mp$0.000 & 0.564$\mp$0.006 & 0.901$\mp$0.013 \\
&$p$-value (vs RW)& 0.088 & 0.763 & 0.043 & 0.096 & 0.002 & 0.047 & 0.095 & 0.459 & 0.101 & 0.000 & 0.249 \\
&$p$-value (vs P)& \textbf{0.582} & 0.091 & 0.086 & 0.467 & 0.129 & 0.084 & 0.700 & 0.632 & 0.535 & 0.000 & 0.016 \\
\hline \rule{0pt}{3ex}
\multirow{3}{*}{C4.5} & AUC & 0.851$\mp$0.024 & 0.712$\mp$0.049 & 0.742$\mp$0.048 & 0.803$\mp$0.075 & 0.613$\mp$0.057 & 0.570$\mp$0.043 & 0.948$\mp$0.032 & \textbf{0.999$\mp$0.001} & 0.503$\mp$0.009 & 0.982$\mp$0.004 & 0.917$\mp$0.012 \\
&$p$-value (vs RW)& 0.008 & 0.747 & 0.015 & 0.429 & 0.084 & 0.270 & 0.090 & \textbf{0.242} & 0.088 & 0.000 & 0.229 \\
&$p$-value (vs P)& 0.039 & 0.358 & 0.634 & 0.734 & 0.819 & 0.508 & 0.137 & 0.533 & 0.535 & 0.100 & 0.020 \\
\hline \rule{0pt}{3ex}
\multirow{3}{*}{S$k$-NN} & AUC & 0.836$\mp$0.022 & 0.735$\mp$0.073 & \textbf{0.877$\mp$0.042} & \textbf{0.969$\mp$0.015} & 0.807$\mp$0.05 & 0.667$\mp$0.04 & 0.942$\mp$0.04 & 0.998$\mp$0.001 & 0.561$\mp$0.024 & 0.98$\mp$0.005 & 0.907$\mp$0.009 \\
&$p$-value (vs RW)& 0.008 & 0.172 & \textbf{0.350} & \textbf{0.402} & 0.673 & 0.735 & 0.330 & 0.458 & 0.424 & 0.000 & 0.056 \\
&$p$-value (vs P)& 0.039 & 0.374 & 0.165 & 0.063 & 0.036 & 0.006 & 0.602 & 0.571 & 0.167 & 0.121 & 0.015 \\
\hline \rule{0pt}{3ex}
\multirow{3}{*}{SSVM} & AUC & \textbf{0.949$\mp$0.014} & 0.632$\mp$0.054 & 0.865$\mp$0.042 & 0.946$\mp$0.025 & 0.800$\mp$0.031 & 0.687$\mp$0.047 & \textbf{0.984$\mp$0.023} & 0.998$\mp$0.001 & 0.525$\mp$0.017 & 0.987$\mp$0.004 & 0.905$\mp$0.011 \\
&$p$-value (vs RW)& 0.106 & 0.008 & 0.664 & 0.477 & 0.667 & 0.312 & 0.012 & 0.324 & 0.164 & 0.000 & 0.176 \\
&$p$-value (vs P)& \textbf{0.592} & 0.031 & 0.273 & 0.120 & 0.017 & 0.055 & \textbf{0.160} & 0.567 & 0.391 & 0.031 & 0.002 \\
\hline \rule{0pt}{3ex}
\multirow{3}{*}{SC4.5} & AUC & 0.867$\mp$0.016 & 0.686$\mp$0.050 & 0.760$\mp$0.075 & 0.872$\mp$0.070 & 0.688$\mp$0.073 & 0.605$\mp$0.060 & 0.942$\mp$0.025 & 0.997$\mp$0.002 & 0.500$\mp$0.009 & 0.982$\mp$0.005 & 0.922$\mp$0.019 \\
&$p$-value (vs RW)& 0.066 & 0.450 & 0.074 & 0.572 & 0.216 & 0.229 & 0.203 & 0.590 & 0.086 & 0.000 & 0.027 \\
&$p$-value (vs P)& 0.141 & 0.165 & 0.559 & 0.116 & 0.472 & 0.334 & 0.616 & 0.642 & 0.509 & 0.088 & 0.005 \\
\hline \rule{0pt}{3ex}
\multirow{3}{*}{E$k$-NN} & AUC & 0.856$\mp$0.025 & \textbf{0.813$\mp$0.028} & \textbf{0.889$\mp$0.043} & 0.964$\mp$0.004 & \textbf{0.857$\mp$0.024} & \textbf{0.755$\mp$0.040} & 0.960$\mp$0.008 & 0.996$\mp$0.004 & \textbf{0.731$\mp$0.041} & 0.955$\mp$0.006 & 0.913$\mp$0.006 \\
&$p$-value (vs RW)& 0.019 & 0.090 & \textbf{0.494} & 0.379 & \textbf{0.267} & \textbf{0.101} & 0.092 & 0.693 & \textbf{0.007} & 0.000 & 0.060 \\
&$p$-value (vs P)& 0.069 & \textbf{0.320} & 0.108 & 0.077 & 0.005 & 0.000 & 0.638 & 0.534 & 0.006 & 0.787 & 0.001 \\
\hline \rule{0pt}{3ex}
\multirow{3}{*}{ESVM} & AUC & 0.948$\mp$0.008 & 0.783$\mp$0.041 & \textbf{0.896$\mp$0.037} & \textbf{0.970$\mp$0.004} & \textbf{0.862$\mp$0.023} & \textbf{0.749$\mp$0.047} & 0.973$\mp$0.019 & 0.998$\mp$0.002 & \textbf{0.744$\mp$0.026} & 0.735$\mp$0.028 & \textbf{0.953$\mp$0.004} \\
&$p$-value (vs RW)& 0.138 & 0.027 & \textbf{0.376} & \textbf{0.336} & \textbf{0.175} & \textbf{0.209} & 0.023 & 0.347 & \textbf{0.028} & 0.001 & \textbf{0.011} \\
&$p$-value (vs P)& 0.669 & 0.454 & 0.060 & 0.074 & 0.004 & 0.004 & 0.312 & 0.546 & 0.000 & 0.000 & 0.001 \\
\hline \rule{0pt}{3ex}
\multirow{3}{*}{EC4.5} & AUC & 0.909$\mp$0.021 & 0.800$\mp$0.031 & 0.858$\mp$0.043 & 0.957$\mp$0.008 & \textbf{0.835$\mp$0.041} & 0.662$\mp$0.066 & 0.963$\mp$0.012 & \textbf{1.000$\mp$0.000} & \textbf{0.694$\mp$0.054} & \textbf{0.990$\mp$0.006} & \textbf{0.959$\mp$0.005} \\
&$p$-value (vs RW)& 0.397 & 0.068 & 0.430 & 0.397 & \textbf{0.494} & 0.794 & 0.101 & \textbf{0.112} & \textbf{0.004} & 0.000 & \textbf{0.008} \\
&$p$-value (vs P)& 0.415 & 0.691 & 0.153 & 0.084 & 0.018 & 0.083 & 0.553 & 0.529 & 0.014 & \textbf{0.017} & 0.001 \\
\hline \rule{0pt}{3ex}
\multirow{3}{*}{C$k$-NN} & AUC & 0.834$\mp$0.024 & \textbf{0.809$\mp$0.023} & 0.874$\mp$0.039 & \textbf{0.972$\mp$0.008} & 0.819$\mp$0.022 & \textbf{0.692$\mp$0.036} & \textbf{0.986$\mp$0.023} & 0.998$\mp$0.001 & 0.583$\mp$0.039 & 0.987$\mp$0.002 & 0.889$\mp$0.010 \\
&$p$-value (vs RW)& 0.008 & 0.110 & 0.660 & \textbf{0.330} & 0.716 & \textbf{0.459} & 0.007 & 0.366 & 0.549 & 0.000 & 0.552 \\
&$p$-value (vs P)& 0.042 & \textbf{0.382} & 0.138 & 0.067 & 0.006 & 0.035 & \textbf{0.030} & 0.573 & 0.013 & 0.032 & 0.008 \\
\hline \rule{0pt}{3ex}
\multirow{3}{*}{CSVM} & AUC & \textbf{0.951$\mp$0.010} & 0.735$\mp$0.062 & 0.835$\mp$0.042 & 0.948$\mp$0.025 & 0.766$\mp$0.045 & 0.690$\mp$0.062 & \textbf{0.983$\mp$0.027} & 0.984$\mp$0.012 & 0.685$\mp$0.057 & 0.590$\mp$0.011 & \textbf{0.953$\mp$0.006} \\
&$p$-value (vs RW)& 0.068 & 0.735 & 0.569 & 0.370 & 0.207 & 0.638 & 0.009 & 0.459 & 0.257 & 0.000 & \textbf{0.016} \\
&$p$-value (vs P)& \textbf{0.532} & 0.164 & 0.408 & 0.164 & 0.054 & 0.140 & \textbf{0.109} & 0.632 & 0.043 & 0.000 & 0.002 \\
\hline \rule{0pt}{3ex}
\multirow{3}{*}{C5.0} & AUC & 0.865$\mp$0.038 & 0.709$\mp$0.052 & 0.802$\mp$0.020 & 0.905$\mp$0.085 & 0.692$\mp$0.066 & 0.610$\mp$0.068 & 0.939$\mp$0.022 & \textbf{1.000$\mp$0.000} & 0.521$\mp$0.037 & \textbf{0.987$\mp$0.006} & 0.935$\mp$0.015 \\
&$p$-value (vs RW)& 0.020 & 0.683 & 0.395 & 0.694 & 0.219 & 0.614 & 0.265 & \textbf{0.112} & 0.132 & 0.000 & 0.027 \\
&$p$-value (vs P)& 0.005 & 0.347 & 0.514 & 0.034 & 0.439 & 0.109 & 0.436 & 0.529 & 0.342 & \textbf{0.144} & 0.002 \\
\hline \rule{0pt}{3ex}
\end{tabular}
\end{sideways}
}
\caption{Average of AUC values of ten folds, and standard deviations, of CCCD, weak and strong classifiers for data sets.
The $p$-values of 5x2 CV $F$-tests show the results of two-way tests comparing both CCCDs with other classifiers.
Some of best performers are given in bold.}
\label{52test}
\end{table}

\section{Summary and Discussion}	

We assess the classification performance of various classifiers such as RW-CCCD, P-CCCD, $k$-NN, SVM and C4.5 classifiers and their variants when class imbalance occurs, and we illustrate the robustness of CCCD classifiers to the class imbalance in data sets. This imbalance often occurs in real life data sets where, in two-class settings, minority class (the class with fewer number of observations) is usually dwarfed by the majority class. Class imbalances hinder the performance of many classification algorithms. We studied the performance of CCCD classifiers under class imbalance problem by first simulating a two-class setting similar to the one used in \citet{devinney2003}. In this setting, the support of one class is entirely embedded in the support of the other. Drawing equal number of observations from both class supports results in an imbalance between two classes with respect to their overlapping region, called \emph{local (or restricted) class imbalance}. This difference in the class sizes was also the case in the example of \citet{devinney2003}, and it is the reason that CCCD classifiers show better results than the $k$-NN classifier. We show that P-CCCD classifiers with lower $\tau$ values tend to perform better than the ones with higher $\tau$ values. This is merely a result of balls with $\tau=0$ representing the local density of the target class points better. Similarly, the RW-CCCD classifiers with lower $e$ values are better when the dimensionality is low and the class sizes are high. This might indicate that the denser the data set in $\mathbb{R}^d$, the less useful the scores $T_x$. However, fully utilizing the scores usually increases the classification performance.
 	
Analysis of both simulated and real data sets indicate that both CCCD classifiers show robustness to the class imbalance problem. We demonstrated this by studying the effects of the class overlapping problem together with the class imbalance problem. In fact, there are studies in the literature focusing on the performance of classification methods when class overlapping and class imbalance problems occur simultaneously \citep{prati2004,denil2010}. Overlapping of classes is an important factor in the classification of imbalanced data sets; that is, it drastically affects the classification performance of most algorithms. When classes are both imbalanced and overlapping, performance of $k$-NN, SVM and C4.5 classifiers deteriorate whereas CCCD classifiers are not affected as severely as these methods. We use two alternatives of C4.5 classifiers where we prune the decision tree in one and do not in the other. It is known for some time that pruning deteriorates the performance of tree classifiers under class imbalance. Moreover, SVM is robust to moderately imbalanced class sizes but demonstrates no robustness in highly imbalanced cases. However, whether the data set is highly or moderately imbalanced, CCCD classifiers seem to preserve their AUC compared to $k$-NN, SVM and C4.5 classifiers. Hence, our study suggests that CCCD classifiers are appealing alternatives when data have class imbalance. In addition, we mention the effect of the individual class sizes on the class imbalance problem \citep{japkowicz2002}. Whatever the ratio between class sizes is, if the minority class has a substantially high number of points, the effect of imbalances between classes tend to diminish.

The classifiers $k$-NN, SVM and C4.5 are referred to as weak classifiers since,
by construction, they are sensitive to imbalances between classes in data sets.
In addition, we consider three distinct families of methods to establish strong classifiers based on weak classifiers,
and compare them with CCCD classifiers. We conduct simulation studies to determine how the classification performance jointly depends on both (global) class imbalance and class overlapping, parameterized as $q$ and $\delta$, respectively. Finally, we apply all these classifiers on several UCI and KEEL data sets. By using the SVDD method of \citet{tax2004}, we estimated the overlapping ratios of all these data sets. We show that CCCD classifiers outperform or perform comparable to $k$-NN, SVM and C4.5 classifiers for some overlapping and imbalance ratios in both simulated and real data sets. In particular, CCCDs are better than SVM classifiers in highly imbalanced cases. The effect of high class imbalance on SVM classifier is also studied in \citet{akbani2004} and \citet{raskutti2004}.
However, when no imbalance occurs between classes, CCCD classifiers usually show either comparable or slightly worse performance than the other classifiers.
As for strong classifiers, we employ the most successful methods from three families of schemes where EasyEnsemble and SMOTE+ENN methods are among them. In our simulation studies, we demonstrated that CCCD classifiers, especially RW-CCCDs, work well compared to these strong classifiers when there are considerable overlap and the high (local) imbalance between classes.
However, these methods are slightly better than RW-CCCDs
as these strong classifiers are the best performing ones among their respective families \citep{batista2004,lopez2013}. Nevertheless, RW-CCCDs have still high performance compared to these classifiers with additional increase in testing speed.

We also investigate the performance of CCCD classifiers under different conditions.
Specifically, in two different experiments, we simulate two classes where
(i) classes are imbalanced but supports are not overlapping (well separated) and
(ii) classes are balanced and supports are overlapping with increasing dimensionality.
P-CCCD classifiers are better than RW-CCCD classifiers in experiment (i).
Both CCCD classifiers mostly outperform $k$-NN and SVM classifiers when classes are imbalanced and not overlapping,
however RW-CCCD classifiers outperform these classifiers only when dimensionality is sufficiently high.
In experiment (ii),
the classification performance of CCCD classifiers slightly degrade
compared to $k$-NN and SVM classifiers, especially with increasing $d$.
Among CCCD classifiers, random walk covers appear to be better when classes are both overlapping and imbalanced, however our results suggest the use of pure covers when classes are imbalanced and well separated (i.e., not overlapping).
In fact, class supports are often overlapping in real life data sets,
hence RW-CCCD classifiers seem to be more appealing in practice.

In practice, classifiers based on CCCD classifiers resemble prototype selection methods. CCCDs balance the class sizes by defining balls that catch surrounding points of the same class, and discard these points from the training set. The resulting data set is composed of the centers of these balls and associated radii which are used in scaled dissimilarity measures. Although, CCCD classifiers remove substantial amount of observations from the majority class, they preserve (most of) the information with the radii. The bigger the radius, the more likely that the balls of CCCD classifiers contain more points.
The radii could be considered as an indicator of the local (probability) density of the target class.
The real advantage of CCCD classifiers are these prototype sets which are of (approximately) minimum cardinality,
although training time and space of P-CCCDs and RW-CCCDs may be considerably high.
However, the number of points in the prototype set is substantially low, and hence testing speed is increased. In some cases, RW-CCCDs provide classifiers with the least testing space complexity. Only the decision tree based classifiers, C4.5 and C5.0, achieve comparable or slightly more reduction to RW-CCCDs. However, with increasing dimensionality, sizes of these trees grow exponentially, making them less appealing than RW-CCCDs in the sense of classification space complexity. Hence, CCCDs preserve important information regarding the data sets while substantially increasing the testing speed.
In literature, many classifiers have been devised to preserve the information on the deleted majority class points, however they are all ensemble based classifiers which substantially increase both training and testing time complexities. In that regard, CCCDs offer a novel approach to this particular problem.
	
\citet{eveland2005} modified RW-CCCD classifiers to increase the speed of the face detection in which imbalances between classes occur naturally. They did only refer to the real life applications which consist of class imbalances. They did not, however, investigate the relationship between class imbalance and overlapping problems as thoroughly as our study does. On the other hand, establishing class covers with Euclidean balls raise the possibility of using different regions (the regions are Euclidean hyperballs around target class points in CCCD) to balance the data and, thus, construct non-parametric classifiers with better classification performance. Along this line, CCCDs can be generalized using \emph{proximity maps} \citep{jaromczyk1992}. For example, \citet{ceyhan2005pcd} defined \emph{proximity catch digraphs} (PCDs) that are generalized versions of CCCDs. \citet{ceyhan2005pcd} has introduced three families of PCDs
and used them to test spatial point patterns of segregation and association \citep[see][]{ceyhan2005,ceyhan2006,ceyhan2007}.
PCDs can also be used to derive new graph-based classifiers which are potentially robust to the class imbalance problem. The study of their properties and performance is a topic of ongoing research by the authors.

\bibliography{cccdref}
\bibliographystyle{apalike}

\end{document}